\documentclass[lettersize,journal]{IEEEtran}
\usepackage{amsmath,amsfonts}
\usepackage[linesnumbered,ruled,vlined]{algorithm2e}
\usepackage{array}
\usepackage[caption=false,font=normalsize,labelfont=sf,textfont=sf]{subfig}
\usepackage{textcomp}
\usepackage{stfloats}
\usepackage{url}
\usepackage{verbatim}
\usepackage{todonotes}
\usepackage{graphicx}
\usepackage{xspace}
\usepackage{microtype}
\usepackage{graphicx}
\usepackage{booktabs}
\usepackage[colorlinks,
            linkcolor=blue,
            anchorcolor=blue,
            citecolor=blue,
            urlcolor=black]{hyperref}
\usepackage{orcidlink}

\usepackage{amssymb}
\usepackage{color}
\usepackage{graphicx}
\usepackage{multirow}
\usepackage{booktabs}
\usepackage{colortbl}
\usepackage{makecell}
\usepackage{bbding}
\usepackage{subfloat}
\usepackage{bbding}
\usepackage[table]{xcolor}
\definecolor{customgray}{HTML}{DEDEE6}
\definecolor{midgray}{rgb}{0.85, 0.85, 0.85}
\usepackage{graphicx}
\usepackage{ulem}
\usepackage{xspace}

\usepackage{soul}
\usepackage{xcolor} 

\makeatletter
\renewcommand{\@cite}[2]{\textcolor{blue}{[{#1\if@tempswa , #2\fi}]}}
\makeatother

\newcommand{\eg}{\textit{e.g.}\xspace}
\newcommand{\ie}{\textit{i.e.}\xspace}

\newcommand{\etal}{\textit{et al.}\xspace}

\begin{document}

\title{EvTexture++: Event-Driven Texture Enhancement for Video Super-Resolution}

\author{Dachun~Kai~\orcidlink{0009-0003-6308-5320},
        Jiayao~Lu~\orcidlink{0009-0009-5721-4556},
        Yueyi~Zhang~\orcidlink{0000-0003-0788-8826},~\IEEEmembership{Member,~IEEE,} 
        and~Xiaoyan~Sun~\orcidlink{0000-0003-3638-5566},~\IEEEmembership{Senior~Member,~IEEE}
        
\thanks{Received 17 August 2025; revised 18 January 2026; accepted 25 January 2026. Date of publication 2 February 2026; date of current version 7 May 2026. This work was supported by the National Natural Science Foundation of China under Grant 62472399, Grant 62021001, and Grant 62032006. Recommended for acceptance by L. Zhang. \textit{(Corresponding author: Xiaoyan Sun.)}}
\thanks{Dachun Kai, Jiayao Lu, and Xiaoyan Sun are with the MOE Key Laboratory of Brain-Inspired Intelligent Perception and Cognition, University of Science and Technology of China, Hefei 230027, China (e-mail: dachunkai@mail.ustc.edu.cn; lujiayao@mail.ustc.edu.cn; sunxiaoyan@ustc.edu.cn).}
\thanks{Yueyi Zhang is with Midea Group, Shanghai 201700, China (e-mail: zhyuey@gmail.com).}
\thanks{Code is available at: \url{https://github.com/DachunKai/EvTexture}.}
\thanks{Digital Object Identifier 10.1109/TPAMI.2026.3660020}
}

\markboth{IEEE Transactions on Pattern Analysis and Machine Intelligence, Vol. 48, No. 6, June 2026}
{Kai \MakeLowercase{\textit{et al.}}: EvTexture++: Event-Driven Texture Enhancement for Video Super-Resolution}

\maketitle

\begin{abstract}
Event-based vision has drawn increasing attention owing to its distinctive properties, including ultra-high temporal resolution and extreme dynamic range. Recent works have introduced it to video super-resolution (VSR) to enhance flow estimation and temporal alignment. In contrast, this paper shifts the focus of event signals from motion refinement to texture enhancement in VSR. We propose EvTexture++, the first event-driven framework dedicated to texture enhancement in VSR. It leverages high-frequency spatiotemporal details from events to improve texture recovery. EvTexture++ incorporates a customized texture enhancement branch, along with an iterative texture enhancement module that progressively exploits high-temporal-resolution event information for texture restoration. This enables gradual refinement of texture regions across iterations, yielding more accurate and detailed high-resolution outputs. Besides intra-frame texture recovery, large motions could degrade inter-frame temporal consistency, particularly in texture regions, leading to texture flickering. To mitigate this, we further exploit the continuous-time motion cues of events to enhance temporal consistency, introducing a temporal texture alignment module that estimates event-guided texture-aware flow for precise inter-frame texture alignment. Moreover, EvTexture++ is designed as a plug-and-play tool to flexibly boost the performance of existing VSR models. Experiments on five datasets demonstrate that EvTexture++ achieves state-of-the-art performance. When integrated into recent VSR models, it yields significant improvements, with gains of up to 1.55 dB in PSNR on the texture-rich Vid4 dataset.
\end{abstract}


\begin{figure}[t!]
    \centering
    \vspace{0.5mm}
    \hspace*{-0.02\linewidth} \includegraphics[width=0.95\columnwidth]{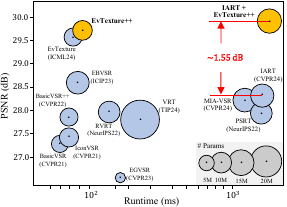}
    \caption{Performance versus runtime on Vid4~\cite{liu2013bayesian}. The circle size corresponds to the number of parameters. Our EvTexture++ achieves a superior trade-off between performance and efficiency. Notably, equipping the state-of-the-art RGB-based VSR model IART~\cite{xu2024enhancing} with our EvTexture++ plug-in yields a remarkable gain of $\sim$1.55~dB. Runtime is measured on $180 \times 320$ LR inputs.}
    \label{fig:fig1}
\end{figure}

\section{Introduction}

\IEEEPARstart{V}{ideo} super-resolution (VSR) aims to reconstruct high-resolution (HR) video frames from low-resolution (LR) inputs, with various applications such as high-definition television~\cite{goto2014super}, video surveillance~\cite{zhang2010super}, virtual reality~\cite{liu2024single}, and video enhancement~\cite{xue2019video}. Compared to single-image super-resolution (SISR), VSR emphasizes modeling inter-frame temporal dependencies, leveraging complementary information from unaligned neighboring frames to restore missing details in the target HR frame.

Over the past few years, VSR approaches have progressed considerably~\cite{xu2024enhancing},~\cite{chan2021basicvsr},~\cite{chan2022basicvsr++},~\cite{hu2023cycmunet+},~\cite{zhou2024video},~\cite{bai2024self}. These approaches typically propagate temporal information via sliding windows~\cite{wang2019edvr} or recurrent networks~\cite{chan2021basicvsr}, with advanced alignment modules (\eg, the flow-guided deformable alignment module~\cite{chan2022basicvsr++}) to handle frame variations. However, RGB-only methods often suffer from insufficient motion information between frames due to the low temporal sampling rates, which hinders accurate temporal alignment and bottlenecks performance.

\begin{figure*}[t!]
        \centering
        \includegraphics[width=\textwidth]{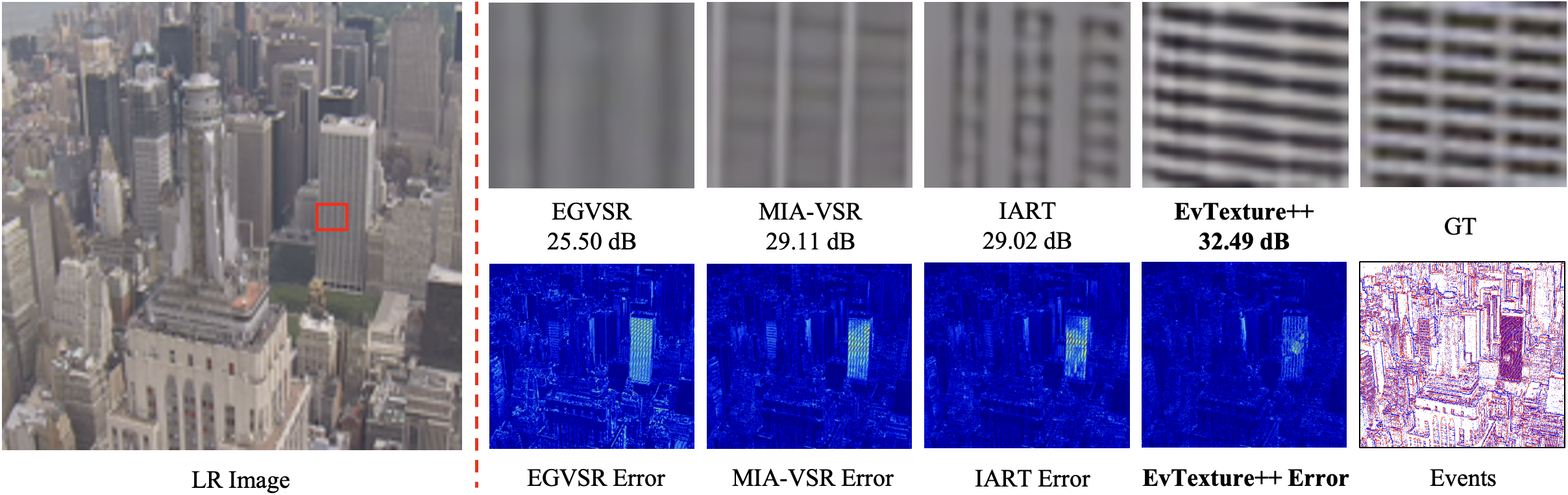}
        \caption{Visual comparison on a challenging texture-rich scene. While current VSR methods, whether frame-based (\eg, MIA-VSR~\cite{zhou2024video}, IART~\cite{xu2024enhancing}) or event-based (\eg, EGVSR~\cite{lu2023learning}), suffer from severe over-smoothing, our EvTexture++ successfully reconstructs coherent building stripes. This is further validated by the error maps (bottom row), where our method exhibits significantly lower residuals by leveraging high-frequency event information.}
        \label{fig:fig2}
    \end{figure*}

More recently, event signals from event cameras~\cite{lichtsteiner2008128} have been introduced to VSR~\cite{jing2021turning},~\cite{lu2023learning},~\cite{kai2023video},~\cite{kai2024evtexture}. Unlike frame-based RGB cameras, event cameras offer ultra-high temporal resolution and wide dynamic range~\cite{gallego2020event}, providing complementary motion cues. For example, EGVSR~\cite{lu2023learning} employs a temporal filtering branch to extract motion features (\eg, edges, corners) from events, while EBVSR~\cite{kai2023video} uses events to refine both optical flow and temporal alignment. Nevertheless, as shown in Fig.~\ref{fig:fig2}, these methods still struggle with fine texture restoration, exhibiting noticeable errors in texture regions.

Texture restoration remains a very challenging and underexplored task in VSR. Recovering HR textural details from LR inputs is inherently difficult, and preserving temporal consistency in texture-rich regions during video playback is equally demanding, given their dense detail. We notice that some efforts have been made on texture enhancement for SISR~\cite{cai2022tdpn},~\cite{liu2024cte}. But in VSR, most methods prioritize addressing issues induced by large motions~\cite{chi2024egovsr} or occluded regions~\cite{chan2022basicvsr++}. To the best of our knowledge, little work has focused specifically on texture restoration in VSR.

We observe that event signals not only offer ultra-high temporal resolution but also contain rich high-frequency dynamic details, making them well-suited for texture restoration in VSR. However, leveraging events for texture enhancement in VSR faces non-trivial challenges: first, events encode relative brightness changes instead of absolute intensity, lacking the baseline intensity context critical for realistic texture reconstruction; second, the asynchronous nature of events leads to spatiotemporal misalignment with synchronous RGB frames, risking mismatched texture hints that could introduce artifacts rather than enhancing details.

To tackle these challenges, we extend our preliminary work EvTexture~\cite{kai2024evtexture} and propose EvTexture++, a more robust event-driven texture enhancement framework. Unlike existing event-based VSR methods that directly use events to estimate HR frames, EvTexture++ progressively recovers high-frequency textural information through two key designs: first, a dedicated texture enhancement branch that explicitly fuses event-derived dynamic cues with RGB intensity information, compensating for the lack of absolute intensity in events; second, an Iterative Texture Enhancement (ITE) module within this branch, which iteratively refines spatiotemporal alignment between events and RGB frames while exploiting sparse event hints in static regions. This refinement across iterations enables more accurate matching of texture details to their spatial locations, yielding HR outputs with both rich dynamics and natural appearance.

Beyond intra-frame texture recovery, large motions often compromise temporal consistency, especially in texture regions. To mitigate this, we further integrate event signals into the motion branch and introduce a Temporal Texture Alignment (TTA) module, which estimates both event-based and RGB-based optical flows to jointly align cross-frame texture regions. The event-based flow is texture-aware, capturing fast, non-linear motions with high temporal precision, while the RGB-based flow provides coarse yet reliable appearance cues. By exploiting the continuous-time motion cues of events and fusing the strengths of both modalities, our model achieves more accurate and stable temporal alignment, reducing texture flickering.

    \begin{figure}[t!]
        \centering
        \includegraphics[width=\columnwidth]{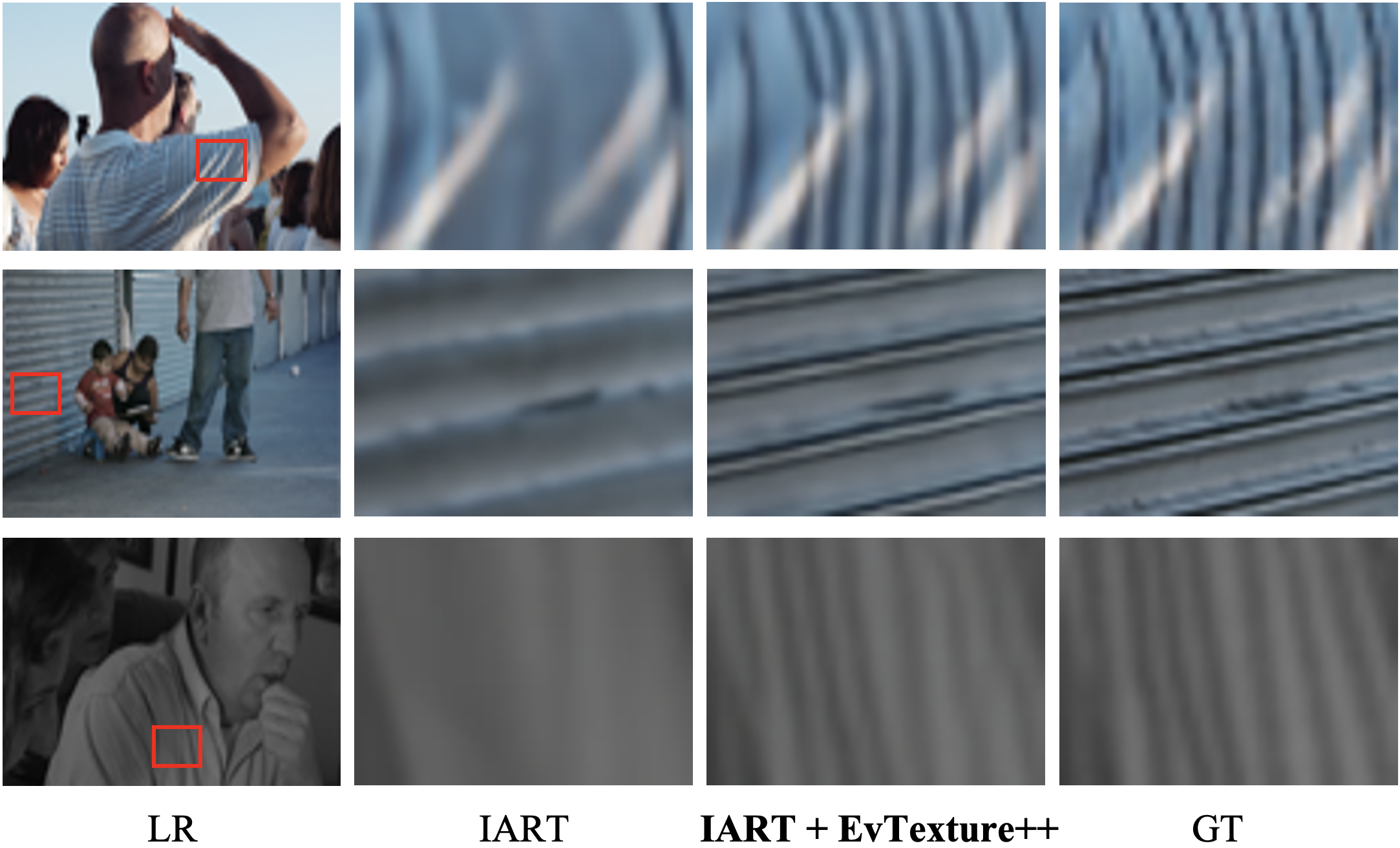}
        \caption{Visual comparison of the plug-in effectiveness on the state-of-the-art VSR model IART~\cite{xu2024enhancing}. While IART tends to over-smooth high-frequency textures, equipping it with our plug-and-play EvTexture++ module successfully recovers these fine structural details, generating results much closer to the GT.}
        \label{fig:fig3}
    \end{figure}

    In our experiments, we first instantiate EvTexture++ on the VSR backbone BasicVSR~\cite{chan2021basicvsr}. The resulting model achieves state-of-the-art (SOTA) performance on various benchmarks (see Fig.~\ref{fig:fig1}). Beyond BasicVSR, advanced VSR backbones, particularly Transformer-based architectures~\cite{xu2024enhancing},~\cite{zhou2024video},~\cite{shi2022rethinking}, have demonstrated impressive performance. However, these models often require substantial training resources and still struggle to restore fine textures. To address these limitations, we designed EvTexture++ as a universal plug-and-play module that seamlessly integrates into diverse architectures, ranging from CNNs to Transformers. During training, the pretrained backbone remains frozen, allowing our module to enhance feature representations with minimal computational overhead. As shown in Fig.~\ref{fig:fig3}, equipping IART~\cite{xu2024enhancing} with our plug-in yields visibly sharper textures. We further apply EvTexture++ to other VSR models, and extensive experiments validate its effectiveness, particularly in recovering fine-grained details. 
    
    Our main contributions can be summarized as follows:
    \begin{itemize}
        \item We present EvTexture++, the first event-driven framework dedicated to texture restoration in VSR.
        \item We propose to progressively recover high-frequency textural information through a dedicated texture enhancement branch coupled with an ITE module.
        \item We introduce a TTA module to enhance temporal texture consistency, which estimates event-guided texture-aware flow for precise inter-frame alignment.
        \item We design EvTexture++ as a flexible plug-and-play tool that can be seamlessly integrated into existing VSR models, significantly improving their performance.
        \item Extensive experiments on five benchmarks show that EvTexture++ achieves new SOTA performance and excels in restoring texture-rich video sequences.
    \end{itemize}

    Note that a preliminary version of this work, EvTexture~\cite{kai2024evtexture}, was accepted to ICML 2024. In this journal version, we not only provide more detailed experiments, evaluations, and discussions, but also present a more robust and flexible framework, EvTexture++, which extends our earlier work in several key aspects:
    (1) In addition to intra-frame texture recovery, EvTexture++ further improves temporal texture consistency under large motion through the TTA module, effectively reducing texture flickering.
    (2) EvTexture++ is designed as a flexible plug-and-play tool that can be seamlessly integrated into existing VSR models to consistently improve performance.
    (3) We conduct a significantly broader experimental evaluation, extending to multiple upsampling scales and additional test sets. Furthermore, we provide a more detailed temporal consistency analysis and a dedicated robustness analysis against large motion and blur-downsampling (BD) degradation.
    (4) Experimental results demonstrate that EvTexture++ and its plug-in variants achieve new SOTA performance on five VSR datasets, with especially strong results on texture-rich videos.

    The rest of this paper is organized as follows. Section~\ref{sec:related-work} reviews the related work.  Section~\ref{sec:method} describes the proposed EvTexture++ framework and its plug-in variants. Section~\ref{sec:experiments} presents extensive experimental results, while Section~\ref{sec:discussion} analyzes the performance under varying levels of texture and motion complexity. Finally, Section~\ref{sec:conclusion} concludes the paper.

\begin{figure*}[t!]
    \centering
    \includegraphics[width=\textwidth]{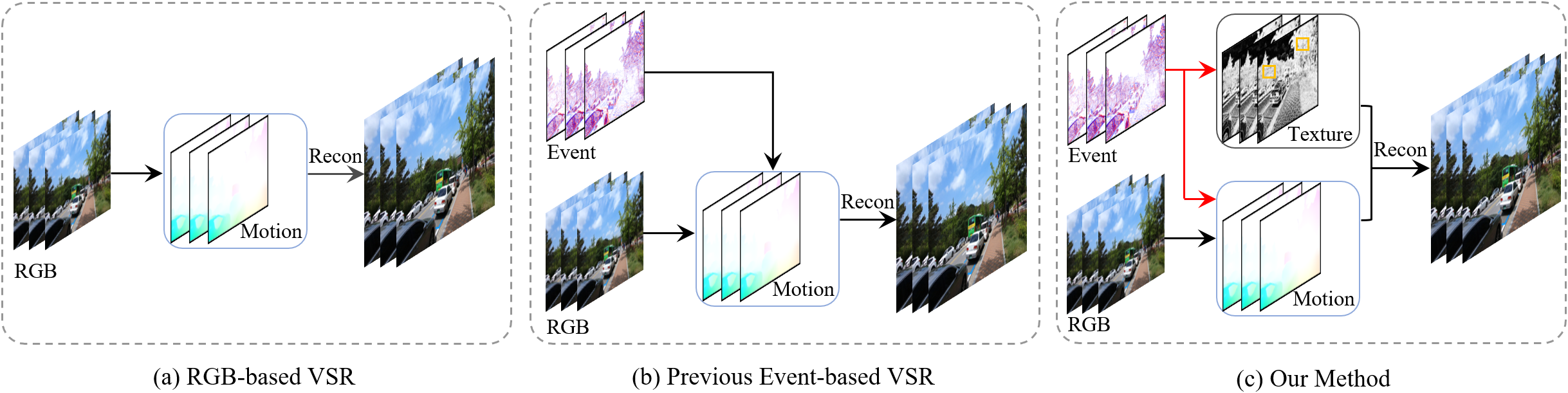}
    \caption{Comparison of different VSR paradigms. (a) RGB-based methods~\cite{xu2024enhancing},~\cite{chan2022basicvsr++},~\cite{zhou2024video} primarily rely on motion alignment to aggregate temporal information. (b) Previous event-based methods~\cite{lu2023learning},~\cite{kai2023video},~\cite{cho2025unifying} leverage events mainly to assist motion learning. (c) In contrast, our method pioneers the use of events for explicit texture restoration, while simultaneously utilizing them to refine motion alignment for better robustness.}
    \label{fig:fig4}
\end{figure*}

\section{Related Work} \label{sec:related-work}

In this section, we review related works on conventional RGB-based VSR methods in Sec.~\ref{related:vsr}, event-based vision in Sec.~\ref{related:event}, and texture restoration in Sec.~\ref{related:texture}.

\subsection{Video Super-Resolution} \label{related:vsr}

In the VSR task, exploiting useful information from other unaligned frames is crucial to recovering missing details in the current frame. Based on how temporal information is utilized, existing VSR methods can be classified into two categories: sliding-window-based and recurrent-based. Sliding-window methods~\cite{wang2019edvr},~\cite{liang2024vrt} use a fixed set of adjacent frames (\eg, 5 or 7 frames) within a temporal window to reconstruct a single HR frame at the center of the window. However, these methods struggle to capture long-range temporal dependencies. To tackle this problem, recurrent structures~\cite{xu2024enhancing},~\cite{chan2021basicvsr},~\cite{chan2022basicvsr++} have become more popular in recent years. These methods use hidden states to propagate information across multiple frames. One of the most well-known models is BasicVSR~\cite{chan2021basicvsr}, which has become a strong, simple, and extensible baseline for VSR. More recently, some VSR models~\cite{xu2024enhancing},~\cite{zhou2024video},~\cite{shi2022rethinking},~\cite{qiu2023learning} have replaced the traditional CNN blocks (\eg, ResNet blocks~\cite{wang2018esrgan}) in BasicVSR with Transformer-based~\cite{vaswani2017attention} blocks, such as those used in SwinIR~\cite{liang2021swinir}, to further improve performance.

In addition to these VSR backbone changes, advanced alignment modules have also been integrated into both sliding-window and recurrent models. Techniques like optical flow estimation~\cite{teed2020raft} and deformable convolutions~\cite{dai2017deformable} are used to create advanced alignment modules. For instance, BasicVSR++~\cite{chan2022basicvsr++} introduced a second-order grid propagation scheme and a flow-guided deformable alignment module to better handle long-term dependencies across misaligned frames. IART~\cite{xu2024enhancing} recently conducted an in-depth analysis of the resampling mechanism and presented an implicit resampling-based alignment method for VSR. However, these methods still face significant challenges when there are large pixel displacements due to the inherent lack of motion information during the blind time between frames.

\begin{figure*}[t!]
    \centering
    \includegraphics[width=\textwidth]{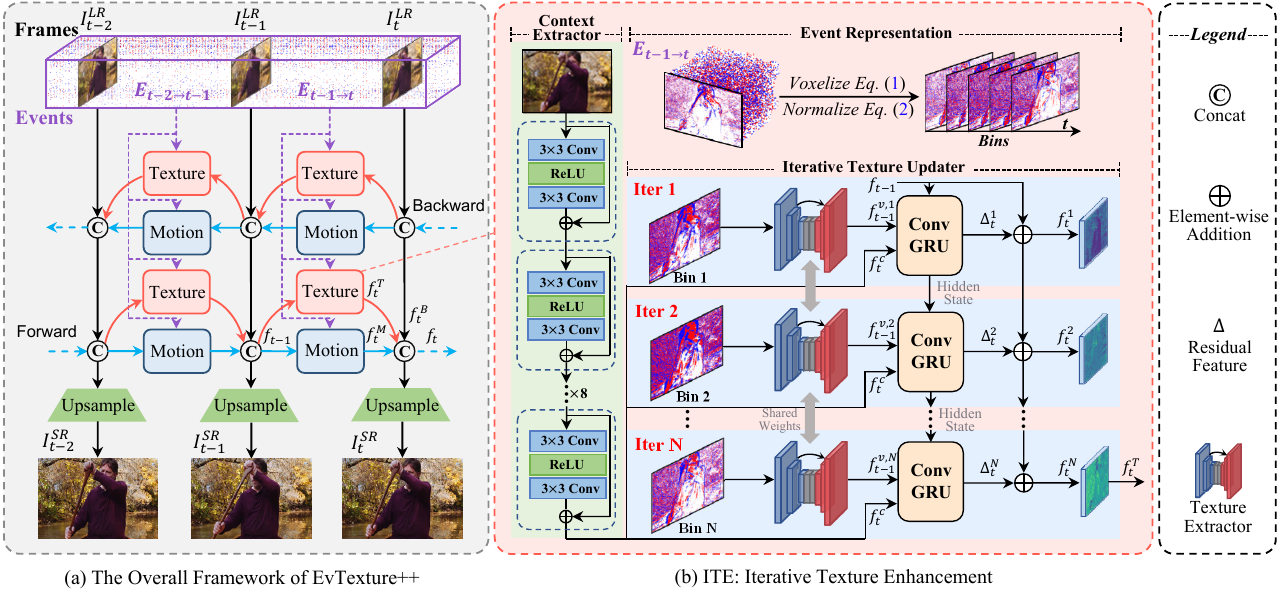}
    \caption{Network architecture of EvTexture++. (a) EvTexture++ adopts a bidirectional recurrent structure with parallel event-guided texture and motion branches for spatial texture restoration and temporal texture consistency, respectively. (b) The ITE module iteratively refines features with richer textural details via a shared ConvGRU, leveraging high-frequency spatiotemporal event signals and the current frame context. The motion branch is illustrated in Fig.~\ref{fig:fig6}.}
    \label{fig:fig5}
\end{figure*}

\subsection{Event-based Vision} \label{related:event}

Event cameras, also known as Dynamic Vision Sensors (DVS)~\cite{lichtsteiner2008128}, are a new type of bio-inspired sensor. Compared to standard RGB cameras, event cameras asynchronously measure per-pixel brightness changes, offering attractive advantages such as ultra-high temporal resolution (around $1~\mu$s), high dynamic range (120 dB), and low power consumption (approximately 5 mW). With these advantages, event cameras have been used in flow estimation~\cite{shiba2024secrets}, frame interpolation~\cite{gao2022superfast}, low-light enhancement~\cite{liu2025ner},~\cite{han2023hybrid}, and motion deblurring~\cite{sun2024unified},~\cite{duan2025eventaid}.

Recently, integrating event signals into VSR has gained increasing attention~\cite{jing2021turning},~\cite{lu2023learning},~\cite{kai2023video},~\cite{xiao2024event},~\cite{kai2025event},~\cite{Xiao_2025_CVPR},~\cite{kai2026seeing}. These methods primarily focus on using events to enhance motion estimation and motion compensation (MEMC), as shown in Fig.~\ref{fig:fig4}(b). For instance, Jing~\etal~\cite{jing2021turning} proposed E-VSR, which first utilizes events to reconstruct intermediate frames. The high-frame-rate video is then encoded into a VSR module to recover HR videos. Kai~\etal~\cite{kai2023video} proposed estimating non-linear flow from events to enhance temporal alignment in VSR. More recently, Xiao~\etal~\cite{Xiao_2025_CVPR} pioneered the application of state space models, particularly Mamba~\cite{zhu2024vision}, to event-driven VSR. Beyond standard VSR, some works~\cite{yan2025evstvsr},~\cite{cho2025unifying} address space-time VSR to simultaneously increase spatial and temporal resolution. Yan~\etal~\cite{yan2025evstvsr} proposed EvSTVSR to specifically tackle large-motion scenarios. However, these methods have not fully exploited the rich high-frequency details of events to address the specific challenge of texture restoration.

\subsection{Texture Restoration} \label{related:texture}

Texture restoration is a fundamental yet understudied issue in VSR. Unlike general SISR, it requires recovering high-frequency details while ensuring temporal consistency during video playback. While texture enhancement has been explored in SISR~\cite{cai2022tdpn},~\cite{liu2024cte},~\cite{fan2025local}, little work has been dedicated to the VSR domain. For example, Cai~\etal~\cite{cai2022tdpn} introduced a texture-learning branch supervised by ground-truth texture components to improve perceptual quality. Recently, Fan~\etal~\cite{fan2025local} proposed a differentiable local texture operator to extract texture structures and introduced a texture branch to predict HR texture distributions. However, to the best of our knowledge, we are the first to dedicate a framework to addressing texture restoration in VSR, leveraging high-frequency event signals to enhance textural details.

\section{Method} \label{sec:method}

In this section, we detail our proposed EvTexture++ framework. First, we begin with the background on event signal representation (Sec.~\ref{sec:event_representation}), followed by an overview of the EvTexture++ framework (Sec.~\ref{sec:framework}). We then describe the core components of EvTexture++, including the event-guided texture branch (Sec.~\ref{sec:texture_branch}), the motion branch (Sec.~\ref{sec:motion_branch}), and the branch fusion module (Sec.~\ref{sec:branch_fusion}). Finally, Section~\ref{sec:plug_in} describes the plug-and-play design of EvTexture++.

\subsection{Event Representation} \label{sec:event_representation}
A raw event stream can be denoted as a set of 4-tuples $E = \{e_k\}_{k=1}^{N_e}$, where $N_e$ represents the number of events. Each event $e_k$ contains four attributes: $ (x_k,y_k,t_k,p_k)$, where $(x_k,y_k)$, $t_k$, and $p_k$ represent the spatial coordinate, timestamp, and polarity of brightness change, respectively.

In practice, it is challenging to effectively represent an asynchronous event stream while preserving complete temporal information. In this work, we adopt the event voxel grid representation $\mathcal{V}$ as in the prior work~\cite{zhu2019unsupervised}, which discretizes the time domain into $B$ time bins. Each time bin in the voxel grid is described as:
\begin{equation}\label{eq1}
    \mathcal{V}(i)=\sum_k p_k \max \left(0,1-\left|\left(i-1\right)-\frac{t_k-t_0}{t_{N_e}-t_{0}}\left(B-1\right)\right|\right),
\end{equation}
where $i \in\{1, \cdots, B\}$ represents the $i$-th time bin. In our experiments, following previous studies~\cite{kai2023video},~\cite{wan2022learning}, we also set $B=5$. Furthermore, to mitigate the impact of hot pixels, we follow the study~\cite{zhu2021eventgan} and normalize the voxel grid $\mathcal{V}$ as:
\begin{equation}\label{eq2}
    \hat{\mathcal{V}}(i)=\frac{\min \left(\mathcal{V}(i), \eta\right)}{\eta},
\end{equation}
where $\eta$ is the 98th percentile value in the non-zero values of $\mathcal{V}$. In this way, we obtain the normalized voxel grid $\hat{\mathcal{V}}\in\mathbb{R}^{H \times W \times B}$, which is amenable to processing by modern deep neural networks. For simplicity, we denote the normalized voxel grid $\hat{\mathcal{V}}$ as $\mathcal{V}$ in the following sections.

\subsection{EvTexture++ Framework} \label{sec:framework}

    We present the EvTexture++ framework, designed to address the challenge of texture restoration in VSR by leveraging high-frequency spatiotemporal details from event signals. The architecture of EvTexture++ is shown in Fig.~\ref{fig:fig5}(a). Specifically, the input is an LR image sequence $\{I^{LR}_t\}_{t=1}^{N_T}$ consisting of $N_T$ frames, along with the corresponding inter-frame events of the $N_T-1$ intervals $\{E_{t\to t+1}\}_{t=1}^{N_T-1}$. The output is the corresponding super-resolved image sequence $\{I^{SR}_t\}_{t=1}^{N_T}$.
	
    Built upon BasicVSR~\cite{chan2021basicvsr}, EvTexture++ adopts a bidirectional recurrent structure, where features are propagated forward and backward, and propagation modules are interconnected. At each timestep, it employs a two-branch structure comprising a texture enhancement branch and a parallel motion learning branch. Crucially, both branches are guided by event signals. The texture branch leverages high-frequency information from events to enhance texture restoration, while the motion branch utilizes events to improve temporal alignment. The outputs of the two branches are fused and then split into two paths. One path is propagated to the next timestep, and the other is upsampled via a pixel-shuffle layer~\cite{shi2016real} and combined with bicubic-upsampled LR frames to generate the final HR output.

    Next, we detail the feature learning process in the texture and motion branches using forward propagation from timestep $t-1$ to $t$ as an example. The only difference in backward propagation lies in the direction of data flow.

\subsection{Event-guided Texture Branch}\label{sec:texture_branch}

    Texture regions are characterized by rich high-frequency details that are difficult to recover. In VSR, this challenge is further exacerbated by the need to maintain temporal consistency across frames. To tackle this, we introduce a dedicated texture enhancement branch designed to restore fine textures using event signals that offer high temporal resolution and capture spatial high-frequency information.
    
    As illustrated in Fig.~\ref{fig:fig5}(b), given the inter-frame event stream $E_{t-1\to t}$ between $I^{LR}_{t-1}$ and $I^{LR}_{t}$, the texture branch takes the propagated feature $f_{t-1}$ as the input state and generates the texture-enhanced feature $f^{T}_{t}$ as:
	\begin{equation}\label{eq4}
		f^{T}_{t}=\mathcal{A}\left(f_{t-1}, E_{t-1\to t}, I^{LR}_{t} \right),
	\end{equation}
    where $\mathcal{A}(\cdot)$ denotes our proposed ITE module. The LR frame $I^{LR}_{t}$ serves as contextual input. By leveraging high-frequency information from events, the texture branch significantly improves the recovery of fine textural details.

    \subsubsection{Iterative Texture Enhancement} \label{sec:ITE}

    Existing methods that directly inject event signals into the HR reconstruction process often overlook the fine-grained temporal structure of events and their correlation with high-frequency textures. To resolve this limitation, we introduce the ITE module, which leverages the high temporal resolution of event streams through voxel-based encoding and iterative refinement. Inspired by the recurrent design of RAFT~\cite{teed2020raft}, our ITE module is designed to model temporal dependencies across voxel bins and progressively enhance texture details. The module consists of two feature extractors and a GRU-based texture updater that iteratively refines the propagated features.

    \textit{Feature Extractors:} We utilize two types of feature extractors: a context extractor $\mathcal{C}$ and a texture extractor $\mathcal{T}$. The parameters of both extractors are shared across all iterations. The context extractor comprises eight residual blocks, as used in ESRGAN~\cite{wang2018esrgan}, and is applied to the current frame $I_t^{LR}$ to obtain the context feature $f^c_t$. The texture extractor is dedicated to extracting the texture feature $f^{v, i}_{t-1}$ from a voxel bin $\mathcal{V}_{t-1\to t}(i)$ at each iteration. It is implemented as a custom five-layer U-Net~\cite{ronneberger2015u}, leveraging its strong ability to capture spatiotemporal features~\cite{jiang2018super}. The feature extraction process can be formulated as:
    \begin{equation}\label{eq5}
        f^c_t=\mathcal{C}\left(I_t^{L R}\right), \quad {f}^{v,i}_{t-1}=\mathcal{T}\left(\mathcal{V}_{t-1\to t}(i)\right).
    \end{equation}
    Here, the context feature $f^c_t$ and the texture feature $f^{v, i}_{t-1}$ match the spatial size $ \mathbb{R}^{C\times H \times W}$ of the propagated feature $f_{t-1}$.

    \begin{figure*}[t!]
        \centering
        \includegraphics[width=\textwidth]{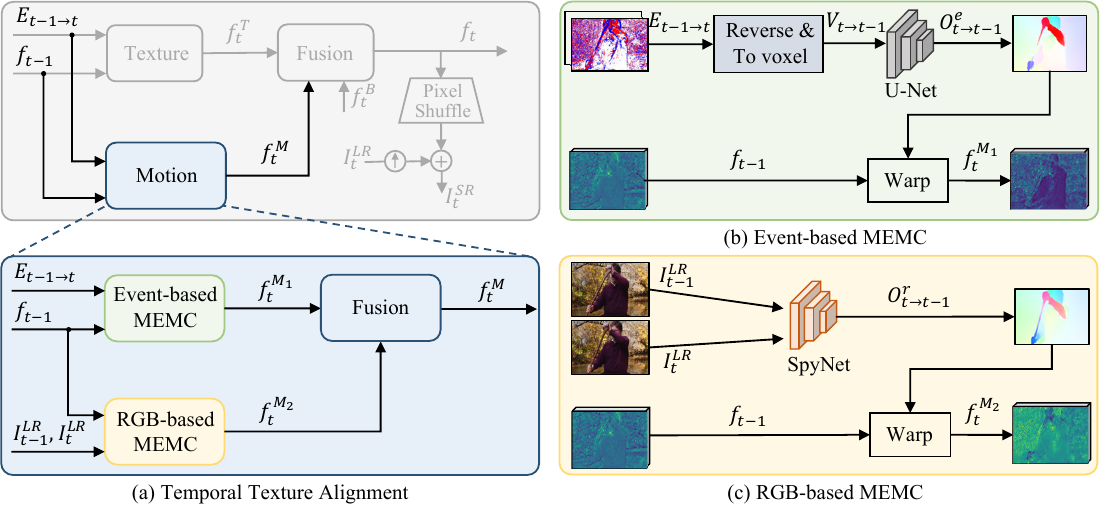}  
        \caption{Event-guided Motion Branch in EvTexture++. (a) EvTexture++ further integrates event signals into the motion branch and introduces a Temporal Texture Alignment (TTA) module, which consists of an RGB-based MEMC and an event-based MEMC that jointly improve feature alignment. MEMC: motion estimation and motion compensation. (b) In the event-based MEMC, events are converted into voxel grids and processed by a U-Net~\cite{jiang2018super} to estimate fast and non-linear motion from events for alignment. (c) The RGB-based MEMC estimates optical flow from images using SpyNet~\cite{ranjan2017optical} and aligns features accordingly.}
        \label{fig:fig6}
    \end{figure*}

    \textit{Iterative Texture Updater:} Subsequent to feature extraction, we introduce a texture updater that fuses textural cues from each voxel bin coupled with context information from the current frame into the propagated feature. The texture updater is shared across iterations and consists of three ConvGRU~\cite{ballas2015delving} layers and five residual blocks~\cite{wang2018esrgan}, denoted as $\mathcal{G}$ and $\mathcal{R}$, respectively. The propagation feature $f_t^i$, initialized with $f_{t-1}$, is updated in a residual manner as:
    \begin{equation}\label{eq6}
        \begin{aligned}
            & h_t^i = \mathcal{G}\left(h^{i-1}_t,\left[f^c_t,f^{v,i}_{t-1}\right]\right), \\ & \Delta^i_t = \mathcal{R}\left(h^i_t\right), \quad f_t^i = f_t^{i-1} + \Delta^i_t,
        \end{aligned}
    \end{equation}
    where $h^i_t$ is the hidden state at timestep $t$ in the $i$-th iteration, with the initial state $h_t^0 = f_{t-1}$. The superscript $i$ denotes the iteration index, where $i \in\{1, \cdots, N\}$. The number of iterations $N$ is set to the number of voxel bins $B$. The operation $[\cdot, \cdot]$ denotes channel-wise concatenation. After $N$ iterations, the final texture-enhanced feature $f_t^{T}$ is obtained as:
    \begin{equation}\label{eq7}
        f_t^{T} = f_{t-1} + \sum\nolimits_{i=1}^{N} \Delta^i_t,
    \end{equation}
    which accumulates the textural refinements contributed by each voxel bin. Notably, $f_t^T$ is obtained via residual learning (by accumulating the learned residuals $\Delta_t^i$ to the input $f_{t-1}$), rather than being directly output by the last ConvGRU layer. The proposed ITE module effectively transfers high-frequency information from event signals to the propagated features through voxel-wise temporal encoding and iterative refinement. By decomposing event streams into temporally discretized voxel bins (as defined in Eqs.~(\ref{eq1}) and~(\ref{eq2})), our method progressively injects fine-grained texture cues over time, thereby facilitating the restoration of complex texture regions.

    \subsection{Event-guided Motion Branch} \label{sec:motion_branch}

    In VSR, large and complex motion often leads to severe misalignment across frames, which significantly degrades reconstruction quality, especially in texture-rich regions. Traditional RGB-based flow estimation is prone to errors under such conditions due to motion blur and limited temporal resolution. To improve robustness, drawing inspiration from prior works on event-based flow estimation~\cite{shiba2024secrets},~\cite{gehrig2024dense} and frame interpolation~\cite{gao2022superfast}, we integrate an event-guided motion branch into EvTexture++. This branch leverages the inherent synergy between event data and RGB frames to achieve more accurate and stable temporal texture alignment.

    \begin{figure*}[t!]
        \centering
        \includegraphics[width=\textwidth]{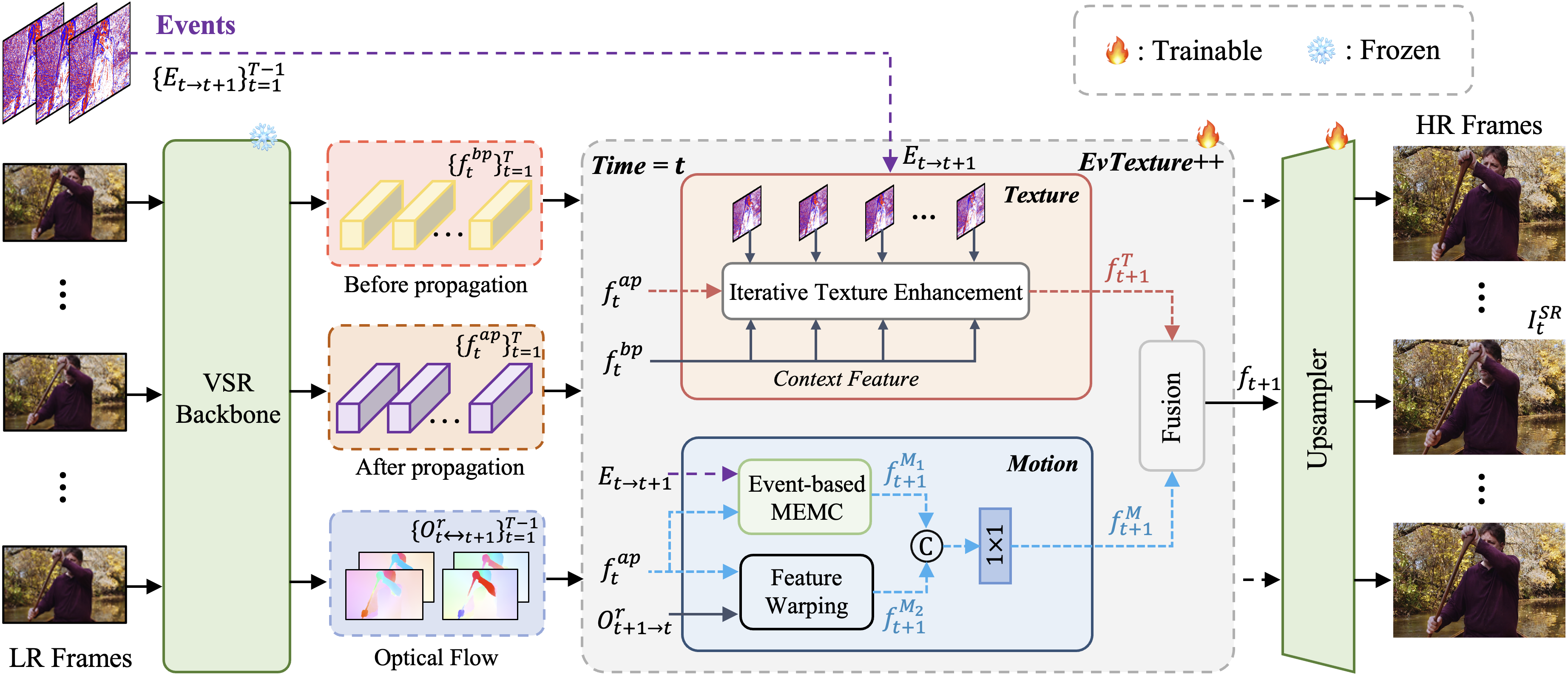}
        \caption{Overview of the EvTexture++ plug-in framework. During training, the frozen backbone extracts three types of features: spatial features before propagation ($f_t^{bp}$), temporal features after propagation ($f_t^{ap}$), and bidirectional optical flow ($O^r_{t\leftrightarrow t+1}$). Our EvTexture++ plug-in refines $f_t^{ap}$ conditioned on event information and the other extracted features. This flexible design can be integrated into various VSR models to consistently improve performance. Note that the motion branch here directly uses the optical flow from the frozen backbone for feature warping and does not require a separate SpyNet (unlike Fig.~\ref{fig:fig6}).}
        \label{fig:fig7}
    \end{figure*}

    As illustrated in Fig.~\ref{fig:fig6}(a), the proposed TTA module plays a crucial role within this branch. The inter-frame events $E_{t-1\to t}$, initially utilized in the texture branch to generate \( f_t^T \), are concurrently exploited by the motion branch. Here, the motion branch propagates \( f_{t-1} \) by integrating event cues with RGB-based flow to produce the motion-enhanced feature \( f_t^M \). These two features ($f_t^T$ and $f_t^M$) are then fused to obtain the final propagated feature at timestep \( t \), denoted as \( f_t \).

    The TTA module comprises a dual-stream structure for motion alignment, consisting of an RGB-based motion estimation and motion compensation (MEMC) branch and an event-based MEMC branch. These two modules work in parallel, focusing on complementary motion cues. The event-based branch exploits events to estimate non-linear motion and align features even in blurred regions, while the RGB-based branch leverages optical flow from frames for feature alignment.
    
    At timestep \( t \), the previous feature \( f_{t-1} \) is aligned to \( f_t^{M_1} \) in the event-based MEMC and to \( f_t^{M_2} \) in the RGB-based MEMC. These two aligned features are then fused to generate the final motion-enhanced feature \( f_t^{M} \). The fusion process can be formulated as:
    \begin{equation}
    f_t^{M} = \mathcal{F_M}([f_t^{M_1}, f_t^{M_2}]),
    \end{equation}
    where \( \mathcal{F_M}(\cdot) \) is a fusion module consisting of feature concatenation followed by a \( 1 \times 1 \) convolution.

    \subsubsection{Event-based MEMC}
    
    As shown in Fig.~\ref{fig:fig6}(b), the event-based MEMC branch first temporally reverses the inter-frame events $E_{t-1\to t}$ and then encodes them into voxel grids following Eqs.~(\ref{eq1}) and~(\ref{eq2}). The voxel grids, denoted as $V_{t\to t-1}$, are processed by a network $\mathcal{U}$ to estimate event-based flow. The network adopts a U-Net architecture, leveraging the design of the event-based frame interpolation method~\cite{tulyakov2021time}. The estimated event-based flow $O_{t \to t-1}^e$ is then used to warp the previous feature $f_{t-1}$, producing the aligned feature $f_t^{M_1}$ from the event-based MEMC branch:
    \begin{equation}\label{eq9}
        O^e_{t\to t-1} = \mathcal{U}(V_{t\to t-1}), \quad f_{t}^{M_1} = \mathcal{W}(f_{t-1}, O^e_{t\to t-1}),
    \end{equation}
    where the superscript $e$ indicates that the flow is derived from events, and $\mathcal{W}(\cdot)$ denotes the backward warping operation. This operation is implemented via differentiable \textit{bilinear interpolation} following the Spatial Transformer Network (STN) mechanism~\cite{jaderberg2015spatial}. Notably, this event-based MEMC branch excels in handling fast or complex motion by exploiting the high temporal resolution of events.

    \subsubsection{RGB-based MEMC}
    Complementary to the event branch, as shown in Fig.~\ref{fig:fig6}(c), we utilize a lightweight optical flow estimation network $\mathcal{S}$ (SpyNet)~\cite{ranjan2017optical} to estimate the optical flow between adjacent frames, following previous VSR studies~\cite{chan2021basicvsr},~\cite{chan2022basicvsr++}. The optical flow is then used to warp the propagated features for temporal alignment. Specifically, we feed $I^{LR}_{t-1}$ and $I^{LR}_{t}$ into the flow estimation network $\mathcal{S}$ to estimate the optical flow $O^r_{t\to t-1}$, where the superscript $r$ denotes that the flow is derived from RGB frames. We then use this flow to backward warp the previous feature \( f_{t-1} \) to the current timestep, resulting in the aligned feature \( f_t^{M_2} \). This process can be formulated as:
    \begin{equation}
    O^r_{t\to t-1}=\mathcal{S}\left(I_t^{L R}, I_{t-1}^{L R}\right), \quad f_{t}^{M_2}=\mathcal{W}\left(f_{t-1}, O^r_{t\to t-1}\right).
    \end{equation}
    This RGB-based MEMC branch captures reliable appearance cues for reconstructing most regions. However, it struggles under large or abrupt motion, where the event-based MEMC branch plays a more important role. Subsequently, the two aligned features, $f_t^{M_1}$ and $f_t^{M_2}$, are integrated to produce the motion-enhanced feature $f_t^M$, enabling more accurate and stable alignment across diverse motion conditions.

    \subsection{Branches Fusion}\label{sec:branch_fusion}
	
    To combine features from the motion and texture branches, we aggregate the motion-enhanced feature $f_t^M$ and the texture-enhanced feature $f_t^T$ using a fusion network. Additionally, the backward-propagated feature $f_t^B$ and the current frame $I_t^{LR}$ are also incorporated to provide contextual guidance. The final propagated feature $f_t$ is computed as:
    \begin{equation}\label{eq8}
        f_t = \mathcal{F}\left(I_t^{LR},f_{t}^{B},\left[f_t^M,f_t^T\right]\right),
    \end{equation}
    where $\mathcal{F}(\cdot)$ denotes the fusion module, comprising fifteen residual blocks~\cite{wang2018esrgan}. The fused feature $f_t$ is then upsampled using a pixel-shuffle layer~\cite{shi2016real}, and the result is combined with the bicubic-upsampled input $I_t^{LR}$ via a residual connection to generate the final high-resolution output $I_t^{SR}$.

    \subsection{EvTexture++ Plug-in} \label{sec:plug_in}    
    Recent SOTA VSR models, particularly Transformer-based methods such as PSRT~\cite{shi2022rethinking}, IART~\cite{xu2024enhancing}, and MIA-VSR~\cite{zhou2024video}, have demonstrated impressive performance. However, these models often demand substantial computational resources. For example, training IART~\cite{xu2024enhancing} on REDS with 8 RTX 4090 GPUs takes approximately 27 days, while MIA-VSR~\cite{zhou2024video} requires about 45 days, based on the settings reported in their original papers.
    Furthermore, these models still struggle to restore fine-grained textures. To address these limitations, we develop EvTexture++ as a flexible plug-and-play module that can be easily integrated into existing VSR models to generally improve their performance, especially in recovering texture regions. The module operates without any modification to the backbone and adds minimal computational overhead. The design and integration of EvTexture++ are detailed below.

    \begin{table*}[t]
	\caption{Quantitative comparison (PSNR$\uparrow$/SSIM$\uparrow$) on Vid4~\cite{liu2013bayesian}, REDS4~\cite{nah2019ntire} and Vimeo-90K-T~\cite{xue2019video} for 4$\times$ VSR. All results are calculated on Y-channel except REDS4 (RGB-channel). \textbf{Bold} and \underline{underlined} values indicate the best and second-best performances.\if Blanked entries correspond to results not reported in previous works.\fi}
    \label{table:table1}
	\centering
        \resizebox{\textwidth}{!}{
		\begin{tabular}{llccccccc}
			\toprule[0.15em]
			& \multirow{2}[2]{*}{Method}  & \multicolumn{5}{c}{Vid4~\cite{liu2013bayesian}} & \multirow{2}[2]{*}{REDS4~\cite{nah2019ntire}} & \multirow{2}[2]{*}{Vimeo-90K-T~\cite{xue2019video}} \\[-0.1em]
			
			\cmidrule(lr){3-7} 
			& & Calendar & City & Foliage & Walk & Average & & \\
			
			\midrule

            \multirow{10}{*}{\makecell{RGB-\\based}} & EDVR~\cite{wang2019edvr} &  23.98/0.8143 & 27.83/0.8112 & 26.34/0.7560  & 31.06/0.9153 & 27.30/0.8242 & 31.09/0.8800 & 37.61/0.9489  \\
			& BasicVSR~\cite{chan2021basicvsr} & 23.87/0.8094 & 27.66/0.8050 & 26.47/0.7710 & 30.96/0.9148 & 27.32/0.8265 & 31.42/0.8909 & 37.18/0.9450 \\
			& IconVSR~\cite{chan2021basicvsr} & 24.07/0.8143  & 27.86/0.8111 & 26.54/0.7705 & 31.08/0.9158 & 27.46/0.8290 & 31.67/0.8948 & 37.47/0.9476  \\
			& RTVAR~\cite{zhou2022revisiting} & 24.65/0.8270 & 29.92/0.8428 & 26.41/0.7652 & 31.15/0.9167 & 27.90/0.8380  & 31.30/0.8850 & 37.84/0.9498\\
			& BasicVSR++~\cite{chan2022basicvsr++} & 24.50/0.8288 & 28.05/0.8212 & 26.90/0.7868 & 31.71/0.9236 & 27.87/0.8413  & 32.39/0.9069 & 37.79/0.9500 \\
            & VRT~\cite{liang2024vrt} & 24.52/0.8296 & 28.33/0.8308 & 26.78/0.7754 & 31.89/0.9258 & 27.88/0.8404   & 32.19/0.9006 & 38.20/0.9530 \\ 
			& RVRT~\cite{liang2022recurrent}  &24.55/0.8334 & 28.35/0.8363 & 26.98/0.7824 & 31.86/0.9251 & 27.94/0.8443 & 32.75/0.9113 & 38.15/0.9527\\
            & PSRT~\cite{shi2022rethinking} & 24.68/0.8357 & 28.71/0.8414 & 27.07/0.7936 & 32.05/0.9289 & 28.20/0.8504 & 32.72/0.9106 & 38.27/0.9536 \\
            & MIA-VSR~\cite{zhou2024video} & 24.48/0.8332 & 29.04/0.8402 & 27.16/0.8008 & 32.17/0.9299 & 28.27/0.8519 & 32.79/0.9115 & 38.22/0.9532 \\
            & IART~\cite{xu2024enhancing} & 24.57/0.8353 & 29.23/0.8434 & 27.09/0.7976 & 32.14/0.9305 & 28.30/0.8523 & \underline{32.90}/0.9138 & 38.14/0.9528 \\
            \midrule
			\multirow{4}{*}{\makecell{Event-\\based}} & EGVSR~\cite{lu2023learning} &21.53/0.6932 & 26.01/0.7068 & 24.33/0.6651 & 27.39/0.8574 & 24.84/0.7330  & 26.87/0.7790 & 34.62/0.9185 \\
			& EBVSR~\cite{kai2023video} & 25.17/0.8548 & 29.30/0.8846 & 27.31/0.8187 & 31.91/0.9265 & 28.46/0.8701 & 31.47/0.8919 & 37.56/0.9490 \\
			& EvTexture~\cite{kai2024evtexture}  & \underline{26.10}/\underline{0.8756} & \underline{31.24}/\underline{0.9087} & \underline{28.12}/\underline{0.8475} & \underline{32.67}/\underline{0.9366} & \underline{29.51}/\underline{0.8909} & 32.79/\underline{0.9174} & \underline{38.23}/\underline{0.9544} \\
			& \textbf{EvTexture++}  & \textbf{26.44}/\textbf{0.8859} & \textbf{31.82}/\textbf{0.9217} & \textbf{28.21}/\textbf{0.8542} & \textbf{32.86}/\textbf{0.9381} & \textbf{29.78}/\textbf{0.8983} & \textbf{32.93}/\textbf{0.9195} &  \textbf{38.32}/\textbf{0.9558} \\
			\bottomrule[0.15em]
		\end{tabular}
         }
\end{table*}

    A typical VSR model usually consists of three stages: (1) frame-wise feature extraction, (2) temporal propagation, and (3) upsampling. The first two stages are handled by the backbone model, which typically comprises multiple Transformer~\cite{liang2021swinir} or CNN blocks~\cite{wang2018esrgan} and demands significant training time and resources. The upsampling layer incurs negligible computational cost. When using our EvTexture++ plug-in, the pretrained backbone remains frozen, and the plug-in is inserted between the backbone and the upsampler to refine the propagated features through our event-guided branches. Only the plug-in and upsampler are updated. Notably, since the backbone is frozen, its internal optical flow estimator is not jointly trained with the plug-in, ensuring training efficiency.

    As shown in Fig.~\ref{fig:fig7}, given an LR input video sequence \( \{I_t^{LR}\}_{t=1}^{T} \), the frozen VSR backbone produces three sets of intermediate outputs: (i) the frame-wise features before propagation \( \{f_t^{bp}\}_{t=1}^{T} \), (ii) the features after propagation \( \{f_t^{ap}\}_{t=1}^{T} \), and (iii) the bidirectional optical flows \( \{O^r_{t \leftrightarrow t+1}\}_{t=1}^{T-1} \). This process is formulated as:
    \begin{equation}
    \{f_t^{bp}\}, \{f_t^{ap}\}, \{O^r_{t \leftrightarrow t+1}\} = \text{VSR-Backbone}(\{I_t^{LR}\}).
    \end{equation}
    The plug-in enhances the propagated feature \( f_{t}^{ap} \) by leveraging inter-frame events \( \{E_{t\to{t+1}}\}_{t=1}^{T-1} \) through the proposed event-guided texture and motion branches. Specifically, for forward propagation from \( t \) to \( t+1 \), the ITE module refines \( f_t^{ap} \) using the contextual feature  \( f_t^{bp} \) and inter-frame events $E_{t\to t+1}$ as:
    \begin{equation}
        f_{t+1}^T = \text{ITE}(f_t^{ap}, f_t^{bp}, E_{t\to t+1}),
    \end{equation}
    where \( f_{t+1}^T \) denotes the texture-enhanced feature with high-frequency details, and ITE is the iterative texture enhancement module introduced in Section~\ref{sec:ITE}.

    In the motion branch, distinct from the standalone architecture which employs a separate SpyNet (Fig.~\ref{fig:fig6}(c)), the plug-in framework avoids redundant computation by directly reusing the pre-computed optical flow from the backbone. Specifically, the event-based flow is estimated from events, while the RGB-based alignment is performed via feature warping using the backbone's flow:
    \begin{equation}
        f_{t+1}^{M_1} = \mathcal{W}(f_{t}^{ap}, O_{t+1 \to t}^e), \quad f_{t+1}^{M_2} = \mathcal{W}(f_{t}^{ap}, O_{t+1 \to t}^r).
    \end{equation}
    Here \( O_{t+1 \to t}^e \) is the event-derived texture-aware flow estimated from voxelized events, and \( O_{t+1 \to t}^r \) is the reused RGB-based optical flow from the frozen backbone. The aligned features are fused to obtain a motion-enhanced representation $f_{t+1}^{M}$. Finally, the texture-enhanced feature $f_{t+1}^T$ and the motion-enhanced feature $f_{t+1}^{M}$ are fused and passed to the upsampler~\cite{shi2016real} to reconstruct the final super-resolved output $I_t^{SR}$.

        \begin{table*}[t]
    \caption{Experiments with different upsampling scales (8$\times$ and 2$\times$) on the REDS4~\cite{nah2019ntire}, UDM10~\cite{yi2019progressive}, Vimeo-90K-T~\cite{xue2019video}, and Vid4~\cite{liu2013bayesian} datasets.}
    \label{table:table2}
    \centering
    \resizebox{\textwidth}{!}{
    \begin{tabular}{lcccccccccccc}
    \toprule[0.15em]
    \multirow{2}[2]{*}{Method} & \multicolumn{3}{c}{\makecell{REDS4~\cite{nah2019ntire} (8$\times$)\\{\scriptsize\textit{(90$\times$160$\to$720$\times$1280)}}}} &  \multicolumn{3}{c}{\makecell{UDM10~\cite{yi2019progressive} (8$\times$)\\{\scriptsize\textit{(90$\times$159$\to$720$\times$1272)}}}} & \multicolumn{3}{c}{\makecell{Vimeo-90K-T~\cite{xue2019video} (2$\times$)\\{\scriptsize \textit{(128$\times$224$\to$256$\times$448)}}}} & \multicolumn{3}{c}{\makecell{Vid4~\cite{liu2013bayesian} (2$\times$)\\{\scriptsize\textit{(240$\times$360$\to$480$\times$720)}}}} \\
    \cmidrule(lr){2-4} \cmidrule(lr){5-7} \cmidrule(lr){8-10} \cmidrule(lr){11-13}
    & PSNR$\uparrow$ & SSIM$\uparrow$ & LPIPS$\downarrow$ & PSNR$\uparrow$ & SSIM$\uparrow$ & LPIPS$\downarrow$ & PSNR$\uparrow$ & SSIM$\uparrow$ & LPIPS$\downarrow$ & PSNR$\uparrow$ & SSIM$\uparrow$ & LPIPS$\downarrow$ \\
    \midrule
    BasicVSR~\cite{chan2021basicvsr} & 26.72 & 0.7475 & 0.3700 & 30.78 & 0.8723 & 0.2532 & 44.69 & 0.9877 & 0.0448 & 34.65 & 0.9598 & 0.0867 \\
    BasicVSR++~\cite{chan2022basicvsr++} & 27.14 & 0.7621 & 0.3549 & 31.36 & 0.8800  & 0.2446 & 45.09 & 0.9880 & 0.0519 & 34.84 & 0.9614 & 0.0829 \\
    EGVSR~\cite{lu2023learning} & 24.12 & 0.6539 & 0.4636 & 27.47 & 0.8047 & 0.3218 & 40.21 & 0.9738 & 0.0699 & 30.42 & 0.9152 & 0.1317 \\
    EBVSR~\cite{kai2023video} & 27.23 & 0.7689 & 0.3522 & 31.73 & 0.8922 & 0.2377  & 44.83 & 0.9881 & 0.0438 & 34.57 & 0.9598 & 0.0842 \\
    EvTexture~\cite{kai2024evtexture} & 28.10 & 0.7996 & 0.3288 & 32.72 & 0.9028 & 0.2206 & 45.25 & 0.9890 & 0.0417 & 35.19 & 0.9659 & 0.0785 \\
        \midrule
    \textbf{EvTexture++} & \textbf{28.20} & \textbf{0.8053} & \textbf{0.3229} & \textbf{32.90} & \textbf{0.9046} & \textbf{0.2168} & \textbf{45.34} & \textbf{0.9895} & \textbf{0.0414} & \textbf{35.29} & \textbf{0.9668} & \textbf{0.0782} \\
    \bottomrule[0.15em]
    \end{tabular}
    }
\end{table*}

\section{Experiments} \label{sec:experiments}

In this section, we present comprehensive experiments to validate the effectiveness of the proposed EvTexture++. We first introduce the datasets (Sec.~\ref{sec:datasets}) and implementation details (Sec.~\ref{sec:implementation}). We then compare EvTexture++ with SOTA VSR methods under various settings (Sec.~\ref{sec:comparisons}), and evaluate its plug-in performance when integrated into different backbones (Sec.~\ref{sec:plugin_performenace}). Finally, Section~\ref{sec:ablation} presents ablation studies to comprehensively assess the impact of each component.

\subsection{Datasets} \label{sec:datasets}

We train and evaluate our method on both synthetic and real-world datasets, comprising a total of five test sets. For synthetic datasets, we follow the widely adopted settings in recent VSR studies~\cite{xu2024enhancing},~\cite{chan2021basicvsr} and utilize two popular benchmark datasets, characterized by diverse motion patterns, for training: Vimeo-90K~\cite{xue2019video} and REDS~\cite{nah2019ntire}. Specifically, models trained on Vimeo-90K are evaluated on Vid4~\cite{liu2013bayesian} and Vimeo-90K-T~\cite{xue2019video}, where we conduct both $2\times$ and $4\times$ VSR experiments and compute PSNR and SSIM metrics on the Y channel. For REDS, we employ REDS4~\cite{nah2019ntire} and UDM10~\cite{yi2019progressive} as the evaluation sets, and perform $4\times$ and $8\times$ VSR, with evaluation on the RGB channel. As the Vimeo-90K, Vid4, REDS, and UDM10 datasets lack real event data, we follow prior event-based VSR works~\cite{jing2021turning},~\cite{kai2023video}, and use the ESIM~\cite{rebecq2018esim} event simulator to generate synthetic events from video frames. The simulated events are converted into voxel grids using Eqs.~(\ref{eq1}) and~(\ref{eq2}). The voxel grids are subsequently downsampled via bicubic interpolation to match the LR input, consistent with the frame downsampling strategy to ensure spatial alignment.

    \begin{table}[t!]
	\caption{Quantitative results on CED~\cite{scheerlinck2019ced} for $2\times$ and $4\times$ VSR. Metrics are calculated on the RGB-channel. $^\dagger$ denotes results are reported in EGVSR~\cite{lu2023learning}.}
	\label{table:table3}
	\centering
	\resizebox*{\linewidth}{!}{
		\begin{tabular}{llcccc}
			\toprule[0.15em]
			& \multirow{2}[1]{*}{Method} & \multicolumn{2}{c}{CED~\cite{scheerlinck2019ced} (2$\times$)} & \multicolumn{2}{c}{CED~\cite{scheerlinck2019ced} (4$\times$)} \\ 
			\cmidrule(lr){3-4} \cmidrule(lr){5-6}
			& & PSNR$\uparrow$ & SSIM$\uparrow$ & PSNR$\uparrow$ & SSIM$\uparrow$ \\
			\midrule
			\multirow{5}[1]{*}{\makecell{RGB-\\based}} & DUF~\cite{jo2018deep}$^\dagger$ & 31.09 & 0.9183 & 28.34 & 0.9081 \\
                & SOF~\cite{wang2020deep}$^\dagger$ & 31.84 & 0.9226 & 27.00 & 0.8050 \\
			& TDAN~\cite{tian2020tdan}$^\dagger$ & 33.74 & 0.9398 & 27.88 & 0.8231\\
			& RBPN~\cite{haris2019recurrent}$^\dagger$ & 36.66 & 0.9754 & 29.80 & 0.8975\\
			& BasicVSR~\cite{chan2021basicvsr} & 39.57 & 0.9778 & 32.93 & 0.9001\\
                \midrule
			\multirow{5}[1]{*}{\makecell{Event-\\based}} & E-VSR~\cite{jing2021turning}$^\dagger$ & 37.32 & 0.9783 & 30.15 & 0.9053\\
			& EGVSR~\cite{lu2023learning}$^\dagger$ & 38.69 & 0.9771 & 31.12 & \textbf{0.9211}\\
			& EBVSR~\cite{kai2023video} & 40.14 & 0.9801 & 33.42 & 0.9075\\
			& EvTexture~\cite{kai2024evtexture} & \underline{40.52} & \underline{0.9813} & \underline{33.68} & 0.9112\\
			& \textbf{EvTexture++} & \textbf{40.57} & \textbf{0.9815} & \textbf{33.71} & \underline{0.9126}\\
			\bottomrule[0.15em]
		\end{tabular}
	}
\end{table}

\begin{table}[t!]
	\caption{Temporal consistency on Vid4~\cite{liu2013bayesian} and REDS4~\cite{nah2019ntire} for $4\times$ VSR.} 
	\label{tab:table4}
	\centering
	\resizebox{\columnwidth}{!}{  
		\begin{tabular}{cccccc}
			\toprule[0.15em]
			\multirow{2}[1]{*}{\makecell{Vid4\\\cite{liu2013bayesian}}} & \multirow{2}[1]{*}{\makecell{BasicVSR\\\cite{chan2021basicvsr}}} & \multirow{2}[1]{*}{\makecell{EGVSR\\\cite{lu2023learning}}} & \multirow{2}[1]{*}{\makecell{EBVSR\\\cite{kai2023video}}} & \multirow{2}[1]{*}{\makecell{EvTexture\\\cite{kai2024evtexture}}} & \multirow{2}[1]{*}{\makecell{\textbf{EvTexture++}}} \\[1.3em]
                \midrule
			TCC$\uparrow$\scalebox{0.5}{$ \times  10$} & 2.736 & 1.707 & 3.336 & \underline{3.655} & \textbf{3.832} \\[0.1em]
			tOF$\downarrow$\scalebox{0.5}{$ \times  10$} & 1.366 & 3.191 & 1.321 & \underline{1.226} & \textbf{1.158} \\
			\midrule \midrule
			\multirow{2}[1]{*}{\makecell{REDS4\\\cite{nah2019ntire}}} & \multirow{2}[1]{*}{\makecell{BasicVSR\\\cite{chan2021basicvsr}}} & \multirow{2}[1]{*}{\makecell{EGVSR\\\cite{lu2023learning}}} & \multirow{2}[1]{*}{\makecell{EBVSR\\\cite{kai2023video}}} & \multirow{2}[1]{*}{\makecell{EvTexture\\\cite{kai2024evtexture}}} & \multirow{2}[1]{*}{\makecell{\textbf{EvTexture++}}} \\[1.3em]
                \midrule
			TCC$\uparrow$\scalebox{0.5}{$ \times  10$} & 4.855 & 3.083 & 4.875 & \underline{5.275} & \textbf{5.321} \\[0.1em]
			tOF$\downarrow$\scalebox{0.5}{$ \times  10$} & 7.496 & 15.692 & 7.505 & \underline{7.153} & \textbf{7.017} \\
			\bottomrule[0.15em]
		\end{tabular}
	}
\end{table}

    \begin{table}[t]
    \caption{Comparison of perceptual similarity (LPIPS$\downarrow$) on Vid4~\cite{liu2013bayesian} for 4$\times$ VSR and computational overheads. FLOPs and runtime are computed on one LR frame with the size of 180$\times$320, using an NVIDIA RTX 3090 GPU.}
    \label{table:table5}
    \centering
    \resizebox{0.93\columnwidth}{!}{
    \begin{tabular}{lcccc}
        \toprule[0.15em]
        \multirow{2}[1]{*}{\makecell{Method}} &  \multirow{2}[1]{*}{\makecell{LPIPS$\downarrow$}} & \multirow{2}[1]{*}{\makecell{\#Params \\ (M)}}  & \multirow{2}[1]{*}{\makecell{Runtime \\ (ms)}} & \multirow{2}[1]{*}{\makecell{FLOPs \\ (G)}}  \\[1.4em]
        \midrule
        EDVR~\cite{wang2019edvr} & 0.2641 & 20.63 & 186.2 & 2018.4\\
        BasicVSR~\cite{chan2021basicvsr} & 0.2783 & 6.29 & \textbf{39.5} & 373.1\\
        IconVSR~\cite{chan2021basicvsr} & 0.2722  & 8.69 & 55.4 & 405.6\\
        BasicVSR++~\cite{chan2022basicvsr++} & 0.2593  & 7.32 & 52.5 & 405.6\\
        PSRT~\cite{shi2022rethinking} & 0.2424  & 13.37 & 1162.3 & 1921.9\\
        MIA-VSR~\cite{zhou2024video} & 0.2441 & 16.60 & 834.5 & 1262.5 \\
        IART~\cite{xu2024enhancing} & 0.2441 & 13.41 & 1204.6 & 1972.7\\
        EGVSR~\cite{lu2023learning} & 0.3351  & \textbf{2.58} & 121.1 & \textbf{159.6} \\
        EBVSR~\cite{kai2023video} & 0.2476  & 12.16 & 68.5 & 641.4 \\
        EvTexture~\cite{kai2024evtexture} & 0.2185 & 8.90 & 93.1 & 805.4 \\
        \midrule
        \textbf{EvTexture++} & \textbf{0.2048}  & 10.15 & 95.0 & 808.6\\
        \bottomrule[0.15em]
    \end{tabular}
    }
\end{table}

To validate our method in real-world scenarios, we follow existing event-based VSR methods~\cite{lu2023learning},~\cite{kai2023video} and use the CED~\cite{scheerlinck2019ced} dataset for training and evaluation. The dataset was captured using a DAVIS346~\cite{brandli2014real} event camera, which outputs spatio-temporally synchronized events and frames at a resolution of $260\times346$. For the train-test split configuration, we follow E-VSR~\cite{jing2021turning} and select 11 clips from the full set of 84 clips as the test set. These clips cover both static and dynamic scenes, as well as indoor and outdoor environments. The remaining clips are used for training. When calculating the metrics, we exclude a boundary of 8 pixels to mitigate boundary artifacts and evaluate on the RGB channel.

\subsection{Implementation Details} \label{sec:implementation}

For training on the REDS~\cite{nah2019ntire} and CED~\cite{scheerlinck2019ced} datasets, we use 15 input frames for each training sample. For Vimeo-90K~\cite{xue2019video}, since each clip contains only 7 frames, we flip the sequence, remove the duplicated center frame (\ie, the 8\textsuperscript{th} frame), and concatenate the remaining frames to form a 13-frame input sequence. We set the mini-batch size to 8 and the input frame size to $64\times64$. We augment the training data with random horizontal and vertical flips. To supervise the training, we utilize the Charbonnier loss~\cite{lai2017deep}, formulated as $\mathcal{L} = \frac{1}{T} \sum_{t=1}^{T} \sqrt{\|I_{t}^{HR} - I_{t}^{SR}\|^2 + \varepsilon^2}$, where $\varepsilon$ is a constant set to $1 \times 10^{-12}$ for numerical stability. We adopt the Adam optimizer~\cite{KingBa15} with $\beta_1{=}0.9$ and $\beta_2{=}0.99$, and schedule the learning rate using Cosine Annealing with Restart~\cite{loshchilov2017sgdr}.

We train our EvTexture++ for 300K iterations on each training dataset. Following previous studies~\cite{xu2024enhancing},~\cite{chan2022basicvsr++}, we employ a pretrained SpyNet~\cite{ranjan2017optical} model to estimate RGB-based optical flow, while all other modules are trained from scratch. For SpyNet, we keep the pretrained model frozen for the first 5K iterations and then train it jointly with the main network, using an initial learning rate of $2.5 \times 10^{-5}$. The initial learning rate for the remaining modules is set to $2 \times 10^{-4}$. The entire training process for each dataset is conducted on four NVIDIA RTX 3090 GPUs and takes approximately 4 days. For the EvTexture++ plug-in variants built upon pretrained VSR models~\cite{xu2024enhancing},~\cite{chan2022basicvsr++},~\cite{zhou2024video},~\cite{shi2022rethinking}, we freeze the VSR backbone parameters and optimize only the plug-in modules. These variants are trained for 200K iterations with an initial learning rate of $1 \times 10^{-4}$. Each plug-in model requires around 6 days of training on four NVIDIA RTX 3090 GPUs.

    \begin{figure*}[t!]
        \centering
        \includegraphics[width=0.98\textwidth]{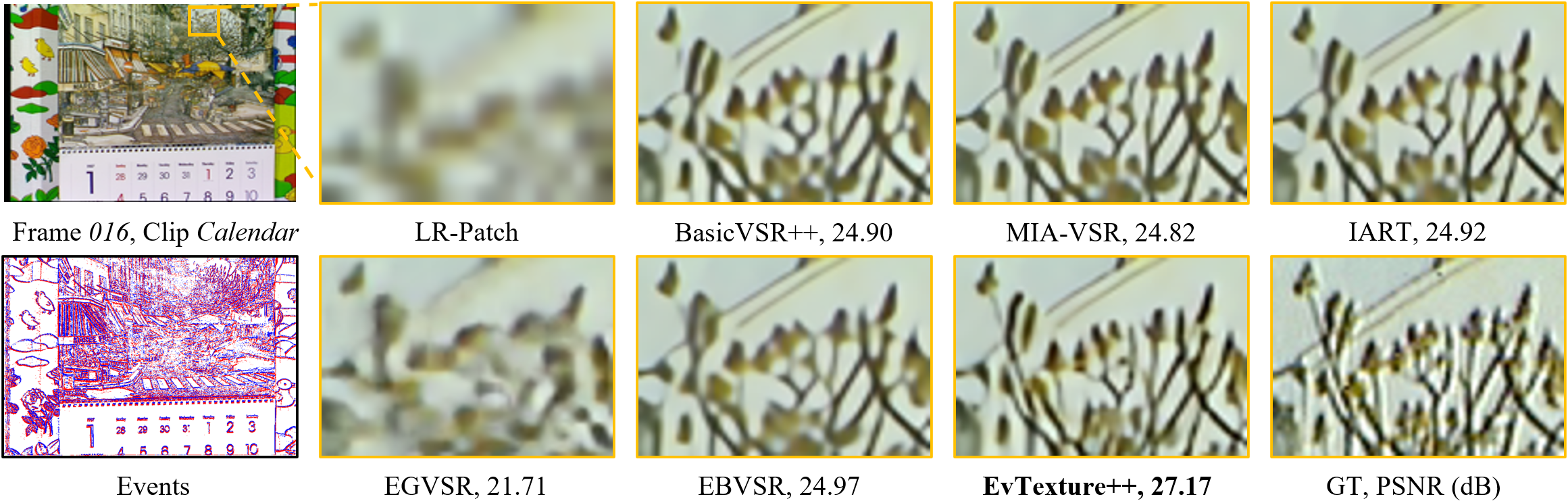}
        \caption{Qualitative comparison on Vid4~\cite{liu2013bayesian} for 4$\times$ VSR. Only EvTexture++ can restore vivid branches and leaves on the tulip tree.}
        \label{fig:fig8}
    \end{figure*}

    \begin{figure*}[t!]
        \centering
        \includegraphics[width=0.98\textwidth]{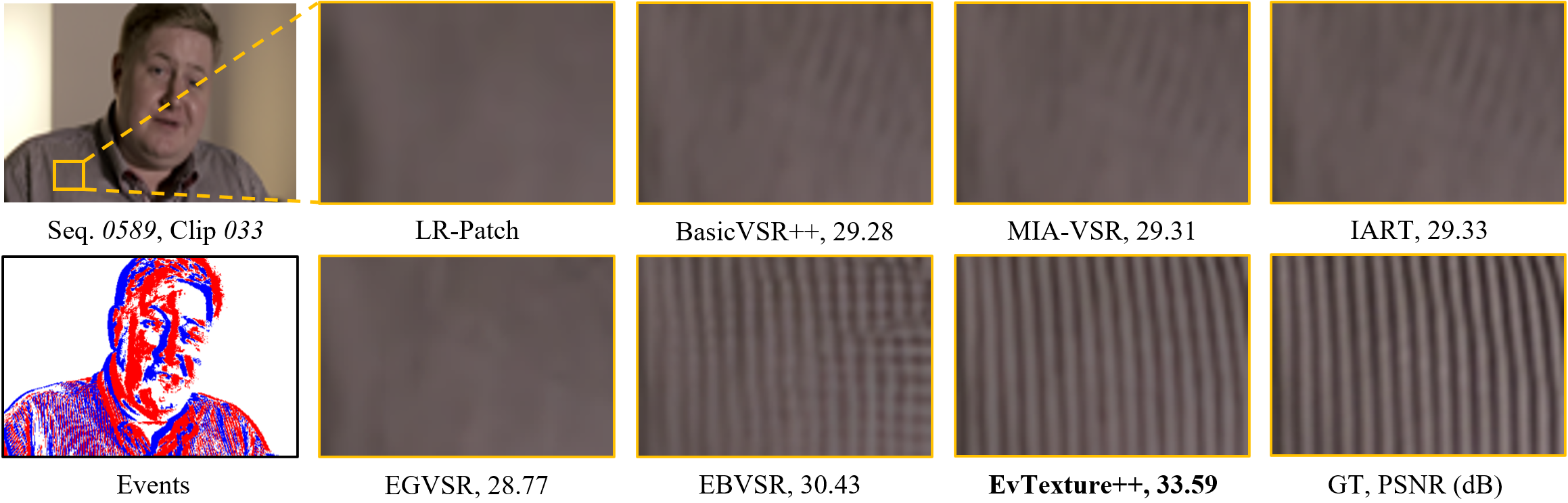}
        \caption{Qualitative comparison on Vimeo-90K-T~\cite{xue2019video} for 4$\times$ VSR. Only EvTexture++ can restore faithful and detailed stripes on clothing surfaces.}
        \label{fig:fig9}
    \end{figure*}

\subsection{Comparisons with State-of-the-Art Methods} \label{sec:comparisons}

\subsubsection{Baselines and Metrics}
    We compare EvTexture++ against two categories of VSR models: (1) RGB-based methods, including BasicVSR~\cite{chan2021basicvsr}, BasicVSR++~\cite{chan2022basicvsr++}, and recent SOTA approaches like MIA-VSR~\cite{zhou2024video} and IART~\cite{xu2024enhancing}; and (2) event-based baselines, comprising EGVSR~\cite{lu2023learning}, EBVSR~\cite{kai2023video}, and our preliminary version, EvTexture~\cite{kai2024evtexture}. To ensure fair comparisons, all methods are trained on the same dataset and evaluated under identical settings. For methods without public code, we report published results. We evaluate spatial quality using PSNR, SSIM, and LPIPS~\cite{blau2018perception}, and assess temporal consistency via tOF~\cite{chu2020learning} and TCC~\cite{chi2020all}, following~\cite{zhang2024tmp}.

\subsubsection{Evaluation on Standard Benchmarks}

Tab.~\ref{table:table1} presents quantitative results on the classic Vid4~\cite{liu2013bayesian}, REDS4~\cite{nah2019ntire}, and Vimeo-90K-T~\cite{xue2019video} datasets. As shown, EvTexture++ significantly outperforms previous SOTA methods. Notably, it surpasses IART~\cite{xu2024enhancing} by 1.48 dB in PSNR on Vid4. Compared to EvTexture~\cite{kai2024evtexture}, EvTexture++ achieves a 0.27 dB gain on Vid4, consistently improving across all clips, as well as an additional 0.16 dB gain on REDS4. Furthermore, as detailed in Tab.~\ref{table:table5}, EvTexture++ attains the lowest LPIPS score of 0.2048 on Vid4, demonstrating superior perceptual quality.

    \begin{figure*}[t!]
        \centering
        \includegraphics[width=0.98\textwidth]{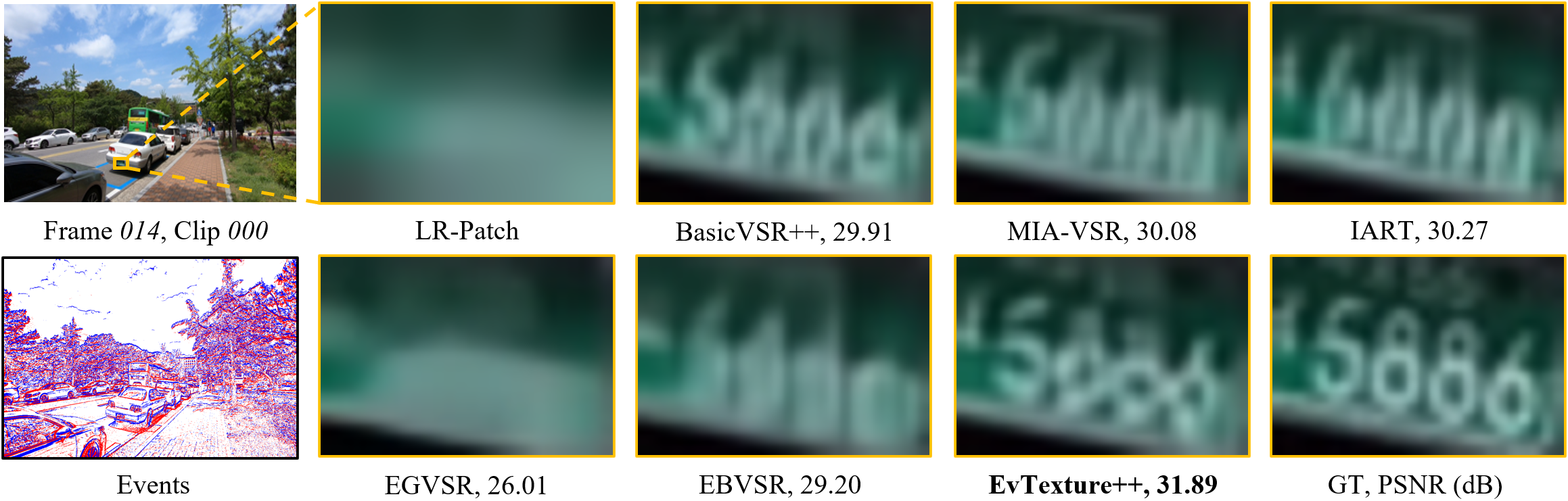}
        \caption{Qualitative comparison on REDS4~\cite{nah2019ntire} for 4$\times$ VSR. Only EvTexture++ can clearly recover the digits ``5886'' on the license plate.}
        \label{fig:fig10}
    \end{figure*}
    
    \begin{figure*}[t!]
        \centering
        \includegraphics[width=0.98\textwidth]{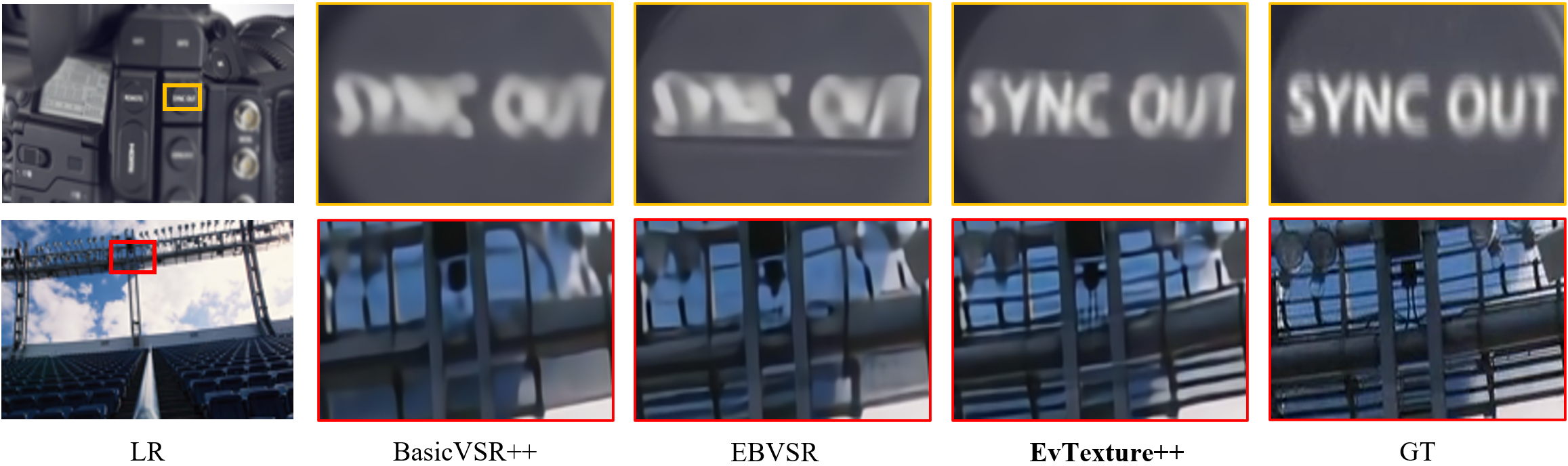}
        \caption{Qualitative comparison on UDM10~\cite{yi2019progressive} for 8$\times$ VSR. Only EvTexture++ can restore fine details such as camera text and railings.}
        \label{fig:fig11}
    \end{figure*}
       
    \begin{figure*}[t!]
        \centering
        \includegraphics[width=0.98\textwidth]{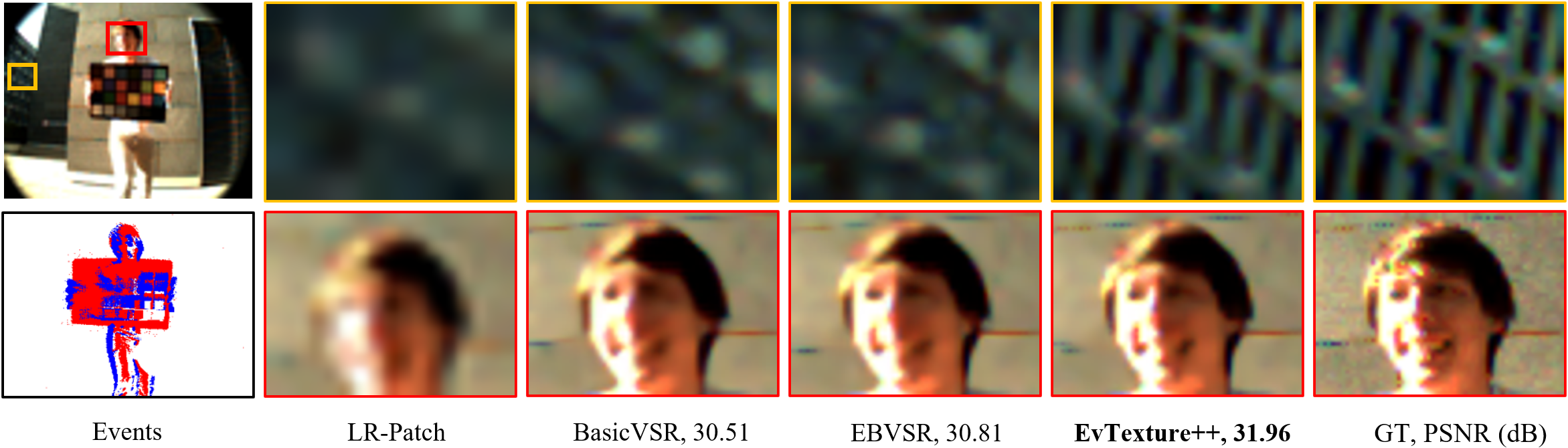}
        \caption{Qualitative comparison on CED~\cite{scheerlinck2019ced} for 4$\times$ VSR. Only EvTexture++ can recover fine wall textures and sharp facial details.}
        \label{fig:fig12}
    \end{figure*}

    \begin{table*}[t]
\caption{Quantitative comparison with our plug-in \colorbox{customgray}{EvTexture++ ($\dagger\dagger$)}. $(*)$ denotes a widened baseline serving as a parameter-equivalent control group. Both $(*)$ and $(\dagger\dagger)$ use a frozen backbone, verifying improvements stem from event cues, not parameter increase.}
\label{table:table6}
\centering
\resizebox{\textwidth}{!}{
    \begin{tabular}{clcccccc}
        \toprule[0.15em]
        & \multirow{2}[1]{*}{Method} & \multirow{2}[1]{*}{REDS4~\cite{nah2019ntire}} & \multirow{2}[1]{*}{Vimeo-90K-T~\cite{xue2019video}} & \multirow{2}[1]{*}{Vid4~\cite{liu2013bayesian}} & \multirow{2}[1]{*}{\#Params (M)}  & \multirow{2}[1]{*}{Runtime (ms)} & \multirow{2}[1]{*}{FLOPs (G)} \\[1.4em]

        \midrule
       & \textcolor{gray}{EvTexture}~\cite{kai2024evtexture} & \textcolor{gray}{32.79/0.9174} & \textcolor{gray}{38.23/0.9544} & \textcolor{gray}{29.51/0.8909} & \textcolor{gray}{8.90} & \textcolor{gray}{93.1} & \textcolor{gray}{805.4} \\
        & \textcolor{gray}{EvTexture++} & \textcolor{gray}{32.93/0.9195} & \textcolor{gray}{38.32/0.9558} & \textcolor{gray}{29.78/0.8983} & \textcolor{gray}{10.15} & \textcolor{gray}{95.0} & \textcolor{gray}{808.6} \\
        
        \midrule
        
        \multirow{3}[1]{*}{\makecell{CNN-based}} & BasicVSR++~\cite{chan2022basicvsr++} &  32.39/0.9069 &  37.79/0.9500 & 27.87/0.8413 & 7.32 & 52.5 & 405.6 \\
        & BasicVSR++$^*$ &  32.29/0.9046 &  37.74/0.9496 & 27.81/0.8387 & 12.97 & 87.4 & 728.3 \\
        & \multicolumn{1}{>{\columncolor{customgray} }l}{\textbf{BasicVSR++$^{\dagger\dagger}$}} &  \multicolumn{1}{>{\columncolor{customgray} }c}{32.94/0.9185} &  \multicolumn{1}{>{\columncolor{customgray} }c}{38.28/0.9547} & \multicolumn{1}{>{\columncolor{customgray} }c}{29.05/0.8806} & \multicolumn{1}{>{\columncolor{customgray} }c}{12.52} & \multicolumn{1}{>{\columncolor{customgray} }c}{118.1} & \multicolumn{1}{>{\columncolor{customgray} }c}{915.7} \\

        \midrule

        \multirow{9}[1]{*}{\makecell{Transformer\\-based}} & PSRT~\cite{shi2022rethinking} &  32.72/0.9138 &  38.27/0.9536 & 28.20/0.8504 & 13.37 & 1162.3 & 1921.9 \\
        & PSRT$^*$ &  32.75/0.9108 &  38.26/0.9535 & 28.19/0.8498 & 19.57 & 1219.6  & 2280.4 \\
        & \multicolumn{1}{>{\columncolor{customgray} }l}{\textbf{PSRT$^{\dagger\dagger}$}} &  \multicolumn{1}{>{\columncolor{customgray} }c}{33.36/0.9233} &  \multicolumn{1}{>{\columncolor{customgray} }c}{38.94/0.9592} & \multicolumn{1}{>{\columncolor{customgray} }c}{29.83/0.8958} & \multicolumn{1}{>{\columncolor{customgray} }c}{19.11} & \multicolumn{1}{>{\columncolor{customgray} }c}{1253.9} & \multicolumn{1}{>{\columncolor{customgray} }c}{2467.8} \\

        \cmidrule{2-8}

         & MIA-VSR~\cite{zhou2024video} &  32.78/0.9220 &  38.22/0.9532 & 28.27/0.8519 & 16.60 & 834.5 & 1262.5 \\
        & MIA-VSR$^*$ &  32.81/0.9119 &  38.32/0.9539 & 28.30/0.8510 & 22.80  & 867.4  & 1788.5  \\
        & \multicolumn{1}{>{\columncolor{customgray} }l}{\textbf{MIA-VSR$^{\dagger\dagger}$}} &  \multicolumn{1}{>{\columncolor{customgray} }c}{33.42/0.9243} &  \multicolumn{1}{>{\columncolor{customgray} }c}{39.02/0.9596} & \multicolumn{1}{>{\columncolor{customgray} }c}{29.81/0.8949} & \multicolumn{1}{>{\columncolor{customgray} }c}{22.34} & \multicolumn{1}{>{\columncolor{customgray} }c}{900.4}  & \multicolumn{1}{>{\columncolor{customgray} }c}{1975.9}  \\

         \cmidrule{2-8}

         & IART~\cite{xu2024enhancing} &  32.90/0.9138 &  38.14/0.9528 & 28.30/0.8523 &  13.41 & 1204.6 & 1972.7\\
        & IART$^*$ &  32.92/0.9142 &  38.24/0.9534 & 28.34/0.8511 & 19.61  & 1273.4  & 2331.2  \\
        & \multicolumn{1}{>{\columncolor{customgray} }l}{\textbf{IART$^{\dagger\dagger}$}} &  \multicolumn{1}{>{\columncolor{customgray} }c}{33.53/0.9261}  &  \multicolumn{1}{>{\columncolor{customgray} }c}{38.91/0.9589} & \multicolumn{1}{>{\columncolor{customgray} }c}{29.85/0.8957} & \multicolumn{1}{>{\columncolor{customgray} }c}{19.16}  & \multicolumn{1}{>{\columncolor{customgray} }c}{1305.0}  & \multicolumn{1}{>{\columncolor{customgray} }c}{2518.6}  \\
    
        \bottomrule[0.15em]
    \end{tabular}
}
\end{table*}

    \begin{figure*}[t!]
        \centering
        \includegraphics[width=\textwidth]{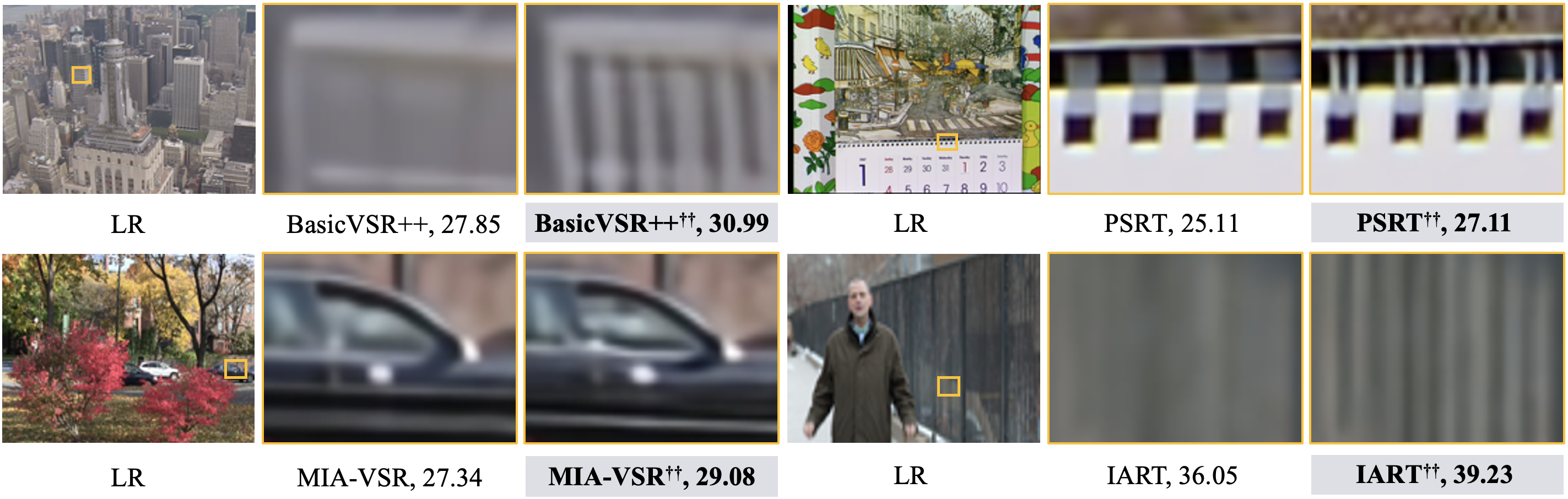}
        \caption{Visual comparison of BasicVSR++\cite{chan2022basicvsr++}, PSRT\cite{shi2022rethinking}, MIA-VSR~\cite{zhou2024video}, and IART~\cite{xu2024enhancing} before and after being equipped with our plug-in \colorbox{customgray}{EvTexture++ ($\dagger\dagger$)}. The PSNR values are indicated in the images. Our EvTexture++ plug-in significantly improves restoration, producing clearer textures and more visual details.}
        \label{fig:fig13}
    \end{figure*}

Qualitative comparisons on Vid4, Vimeo-90K-T, and REDS4 are shown in Figs.~\ref{fig:fig8},~\ref{fig:fig9}, and~\ref{fig:fig10}, respectively. As observed, RGB-based methods such as BasicVSR++~\cite{chan2022basicvsr++} and MIA-VSR~\cite{zhou2024video} struggle to recover fine textures due to the absence of high-frequency information in LR inputs. Prior event-based approaches also exhibit limitations, as they fail to fully leverage the rich high-frequency cues inherent in event signals. In contrast, EvTexture++ consistently produces sharper edges and superior texture details than all competing methods.

\subsubsection{Extended Scale Evaluations}

We also extend our evaluation to different upscaling factors. Specifically, we conduct 8$\times$ VSR on the HR datasets REDS4~\cite{nah2019ntire} and UDM10~\cite{yi2019progressive}, and 2$\times$ VSR on Vid4~\cite{liu2013bayesian} and Vimeo-90K-T~\cite{xue2019video}. The results are summarized in Tab.~\ref{table:table2}. Across all datasets and scales, EvTexture++ consistently achieves the best performance. On REDS4 and UDM10, it outperforms BasicVSR++~\cite{chan2022basicvsr++} by 1.06 dB and 1.54 dB in PSNR, respectively. This confirms the benefit of event information in large-scale upsampling, where RGB frames alone struggle with severe detail loss. EvTexture++ also delivers notable gains on Vimeo-90K-T and Vid4, especially in SSIM and LPIPS, indicating better structure and perceptual quality. Fig.~\ref{fig:fig11} shows qualitative results on UDM10. Our method reconstructs sharper railings and more legible text in the camera, details that are mostly lost in competing methods.

\subsubsection{Evaluation on Real-World Events}

To validate our method on real-world data, we conduct experiments on the CED dataset~\cite{scheerlinck2019ced} under both 2$\times$ and 4$\times$ VSR settings. Quantitative results are summarized in Tab.~\ref{table:table3}. Our EvTexture++ consistently outperforms RGB-based and event-based methods. At the 2$\times$ scale, it achieves 40.57 dB PSNR and 0.9815 SSIM, outperforming the RGB-based competitor, BasicVSR~\cite{chan2021basicvsr}, by 1.00 dB. Compared to the best event-based alternative, EBVSR~\cite{kai2023video}, EvTexture++ further improves performance by 0.43 dB. At the more challenging 4$\times$ scale, EvTexture++ outperforms EBVSR by 0.29 dB in PSNR while maintaining highly competitive SSIM. Qualitative comparisons on CED are shown in Fig.~\ref{fig:fig12}. Our method recovers fine wall textures and sharper facial details, whereas competing methods often produce overly smooth or blurred outputs.

\subsubsection{Temporal Consistency Analysis}

Beyond spatial fidelity, we further evaluate the temporal consistency of super-resolved videos both quantitatively and qualitatively. Tab.~\ref{tab:table4} presents the results on Vid4~\cite{liu2013bayesian} and REDS4~\cite{nah2019ntire} under the 4$\times$ VSR setting. EvTexture++ demonstrates superior temporal consistency on both datasets, yielding lower tOF~\cite{chu2020learning} and higher TCC~\cite{chi2020all} scores than all other methods. 

\begin{figure}[t!]
        \centering
        \hspace*{-0.05\linewidth}
        \includegraphics[width=0.95\columnwidth]{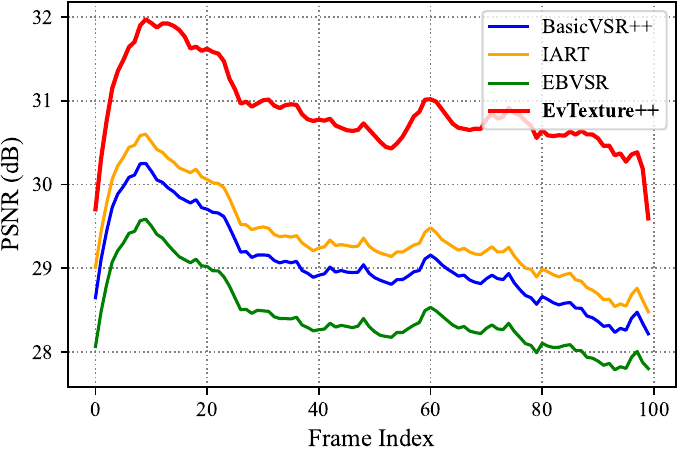}
        \caption{Comparison of per-frame PSNR for `Clip\_000' on REDS4~\cite{nah2019ntire}. Our EvTexture++ achieves higher PSNR than all other methods on every frame, showing better video quality and more stable performance over time.} 
        \label{fig:fig14}
    \end{figure}

\begin{figure}[t!]
        \centering
        \includegraphics[width=\columnwidth]{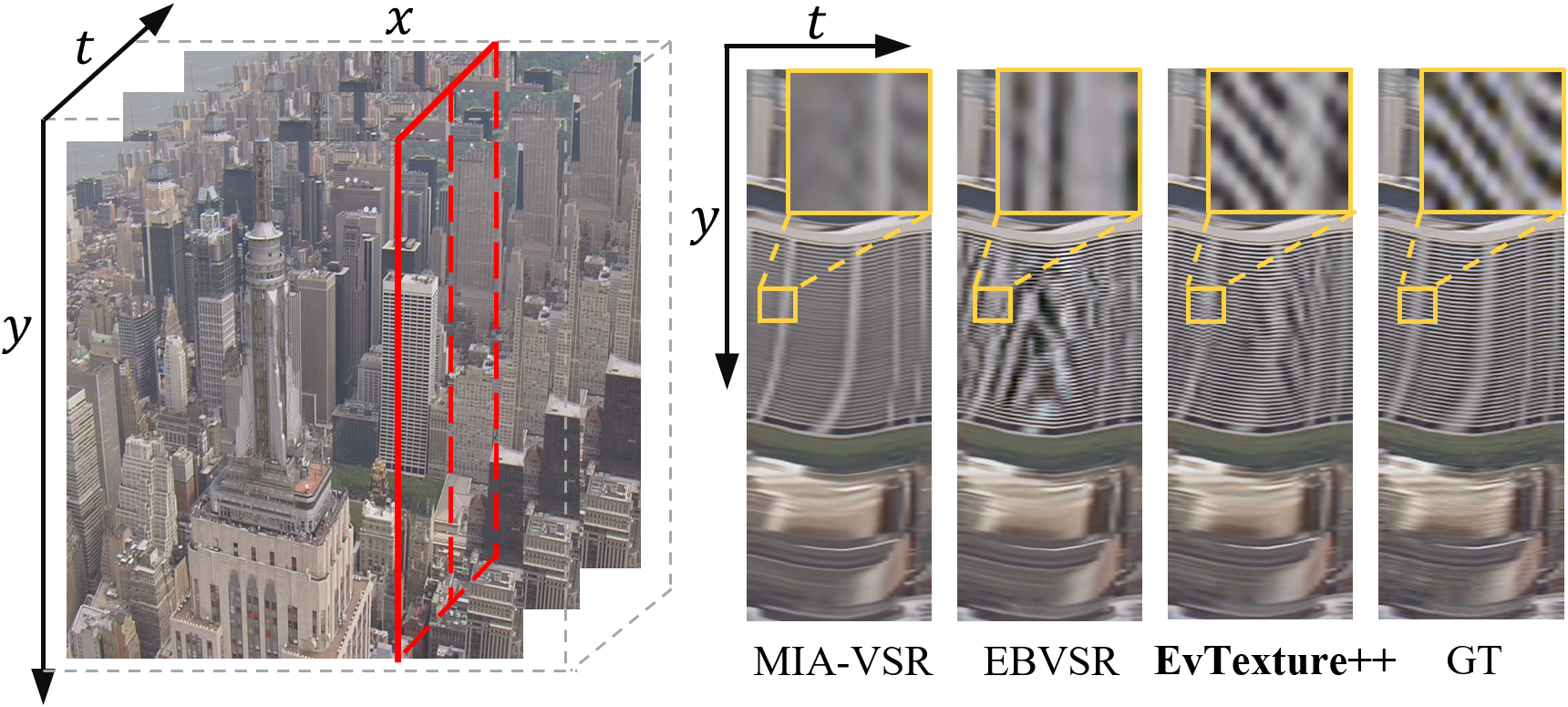}
        \caption{Comparison of temporal profiles. We track a fixed column over time. MIA-VSR~\cite{zhou2024video} and EBVSR~\cite{kai2023video} exhibit flickering artifacts and texture loss. EvTexture++ produces a temporally coherent profile with finer textures.}
        \label{fig:fig15}
    \end{figure}

To further illustrate stability, Fig.~\ref{fig:fig14} shows the per-frame PSNR on `Clip\_000' from REDS4, which contains complex textures. EvTexture++ outperforms all other methods on every frame, indicating not only better video quality but also more stable performance over time. Finally, for qualitative comparisons, we follow the visualization strategy in~\cite{chan2022basicvsr++} and present temporal profiles in Fig.~\ref{fig:fig15}, which visualize temporal transitions by stacking pixel rows over time. As shown, our method exhibits superior consistency in texture regions, which we attribute to the effective utilization of high-temporal-resolution event signals.

\subsubsection{Computational Efficiency}

We evaluate computational efficiency via parameter count, runtime, and FLOPs, as detailed in Tab.~\ref{table:table5}. EvTexture++ employs 10.15 million parameters and requires 808.6 GFLOPs per frame, striking a favorable trade-off between efficiency and performance. It significantly outperforms transformer-based methods like IART~\cite{xu2024enhancing} and MIA-VSR~\cite{zhou2024video}, substantially more efficient, whereas the latter models suffer from per-frame latency exceeding 1000 ms. Compared to EBVSR~\cite{kai2023video} (12.16M parameters), EvTexture++ achieves superior performance with lower computational cost.

\subsubsection{Versatility as a Plug-and-Play Module} \label{sec:plugin_performenace}

We further evaluate the efficacy of our plug-in by integrating it into four representative VSR networks: the CNN-based BasicVSR++~\cite{chan2022basicvsr++} and transformer-based PSRT~\cite{shi2022rethinking}, IART~\cite{xu2024enhancing}, and MIA-VSR~\cite{zhou2024video}. Tab.~\ref{table:table6} demonstrates that our plug-in brings consistent improvements across all baselines. Integrating it into BasicVSR++ boosts PSNR on REDS4~\cite{nah2019ntire} by 0.55 dB with marginal parameter and FLOP overhead. Similar improvements are observed on Vimeo-90K-T~\cite{xue2019video} and Vid4~\cite{liu2013bayesian}. For Transformer-based models, the gains are particularly pronounced. With EvTexture++ integrated, PSRT, MIA-VSR, and IART achieve PSNR gains of 1.63 dB, 1.54 dB, and 1.55 dB, respectively, on the Vid4~\cite{liu2013bayesian} dataset.

To validate that the performance gains of our plug-in stem from the effective utilization of event signals rather than simply from increased model capacity, we conduct control experiments by increasing the parameter counts of the base models to match those of their EvTexture++ counterparts. Specifically, we apply the same ``frozen backbone setting'' to these baselines by appending extra ResNet blocks. As shown in Tab.~\ref{table:table6}, these models, denoted with a superscript $*$, exhibit negligible improvement or even slight degradation. This indicates that naively adding parameters to a frozen backbone without new information leads to redundant computation. This confirms that the observed gains are attributed to the complementary event information leveraged by EvTexture++.

\begin{figure}[t!]
        \centering
        \hspace*{-0.05\linewidth}
        \includegraphics[width=0.96\columnwidth]{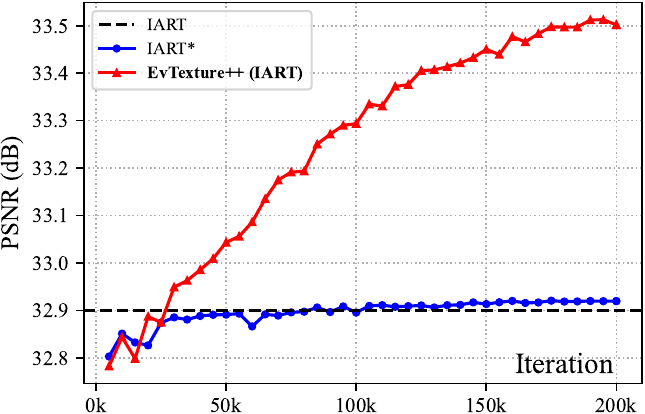}
        \caption{PSNR during training on REDS4~\cite{nah2019ntire} for 4$\times$ VSR using IART-based plug-in variants. IART~\cite{xu2024enhancing} achieves 32.90 dB as reference. $^*$ shows similar parameters as EvTexture++. EvTexture++ significantly improves performance.}
        \label{fig:fig16}
    \end{figure}

Fig.~\ref{fig:fig16} presents the training curves of IART with and without our EvTexture++ plug-in. Notably, the parameter-matched baseline IART$^*$ stagnates, whereas our plug-in achieves superior convergence and performance.
These results demonstrate that EvTexture++ effectively enhances texture quality and serves as a practical, plug-and-play solution compatible with a wide range of VSR models. Fig.~\ref{fig:fig13} show the visual improvements brought by our plug-in. While the base models struggle to recover fine textural details, integrating our EvTexture++ plug-in significantly refines these regions.

\subsection{Ablation Study} \label{sec:ablation}
\label{subsec:abla}

In this section, we first evaluate the impact of incorporating event signals into the texture and motion branches. We then investigate the internal design of the ITE module.

\subsubsection{Effect of Texture and Motion Branches}

\begin{figure}[t!]
        \centering
        \includegraphics[width=\columnwidth]{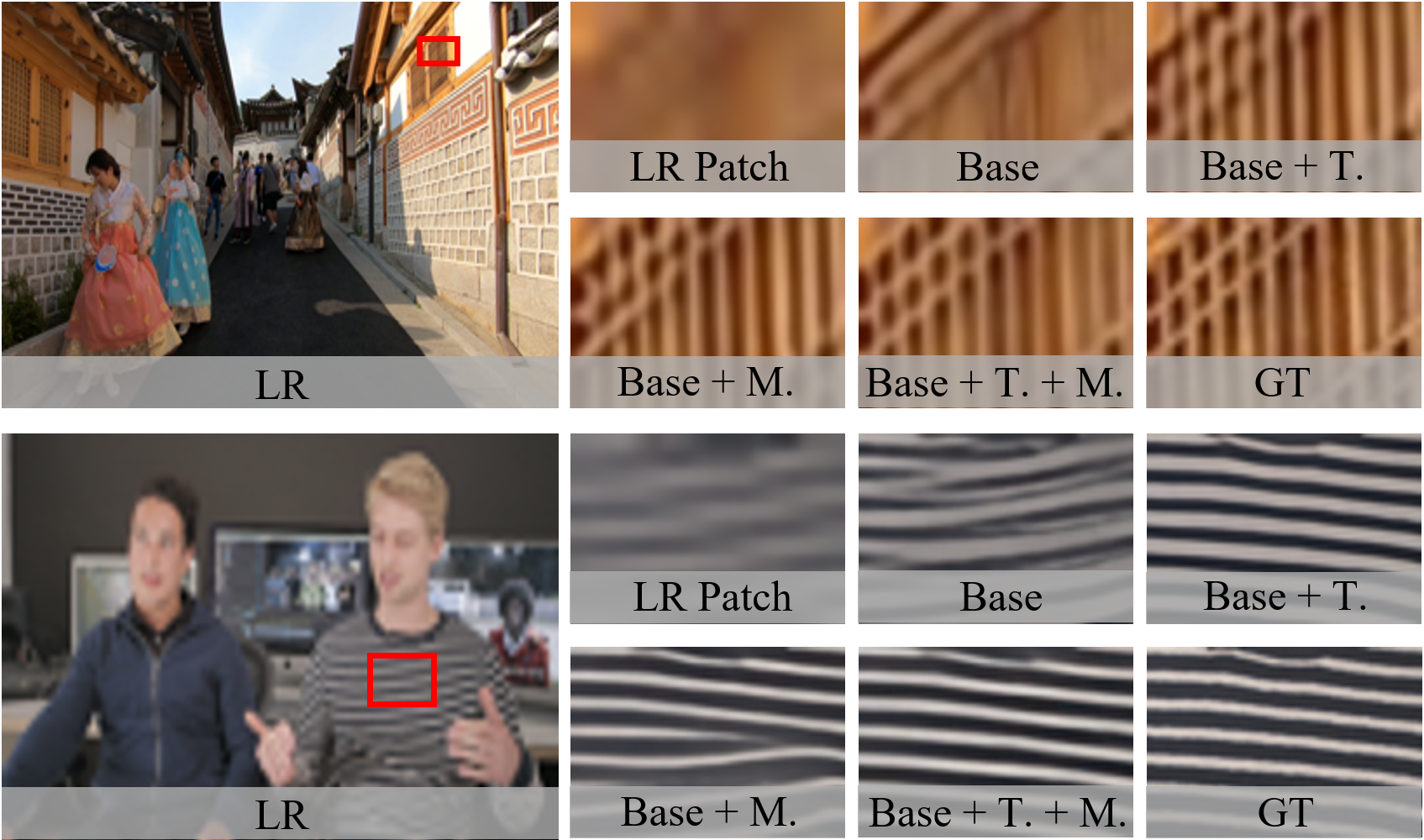}
        \caption{Visual results of ``Base + Texture'', ``Base + Motion'' and ``Base + Texture + Motion'' corresponding to Tab.~\ref{table:table7}. The proposed event-guided texture and motion branches effectively improve reconstruction quality.}
        \label{fig:fig17}
    \end{figure}

We assess the contributions of the event-guided texture and motion branches on Vid4~\cite{liu2013bayesian} and REDS4~\cite{nah2019ntire} for 4$\times$ VSR, with results in Tab.~\ref{table:table7}. The base model operates lacks event information. Introducing the texture branch significantly boosts performance, increasing PSNR from 27.44 dB to 29.51 dB on Vid4 and from 31.58 dB to 32.79 dB on REDS4. These gains validate the effectiveness of event signals in recovering fine textures absent in RGB-based inputs. Solely adding the motion branch also yields improvements, particularly on REDS4, which benefits significantly due to its complex motion. When both branches are employed, the model achieves the best results across both datasets, confirming their complementarity: the texture branch enhances local details, while the motion branch improves dynamic alignment. To visualize these improvements, Fig.~\ref{fig:fig17} presents qualitative comparisons corresponding to Tab.~\ref{table:table7}. The visual results demonstrate that the proposed event-guided modules significantly enhance quality, recovering sharper edges and finer details blurred in the base model.

\begin{figure}[t!]
        \centering
        \includegraphics[width=\columnwidth]{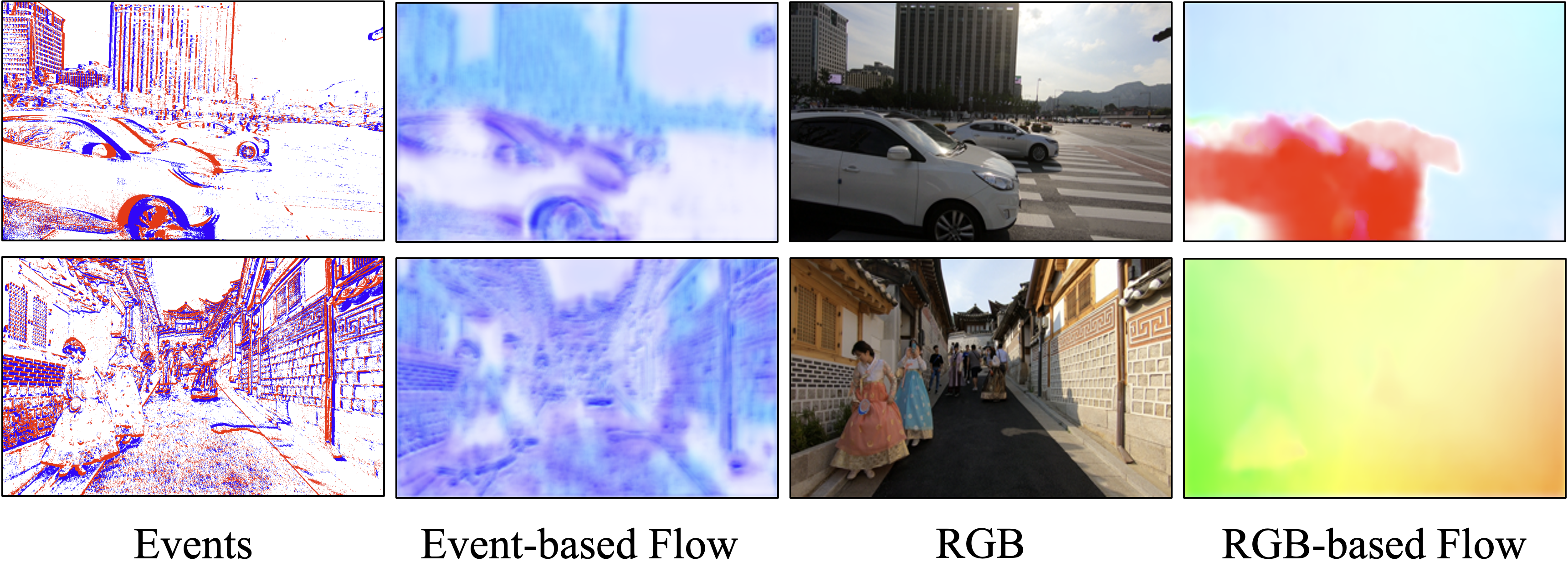}
        \caption{Visual analysis of learned optical flows. Compared to the smooth RGB flow (from the BasicVSR++~\cite{chan2022basicvsr++} backbone), our event-based flow (from the Event-based MEMC module) exhibits sharper boundaries and richer textural details, validating the effective capture of high-frequency motion cues.}
        \label{fig:fig18}
    \end{figure}

To further investigate the mechanism behind the motion branch, we visualize the learned optical flows in Fig.~\ref{fig:fig18}. As shown, even without explicit supervision, the Event-based MEMC module captures meaningful motion structures. Crucially, unlike the smooth RGB flow from the backbone, our event-based flow preserves sharper motion boundaries and richer textural details. This confirms that the module effectively exploits high-frequency event signals to provide complementary alignment cues, explaining the performance gains observed in the motion-only ablation.

    \begin{table}[t!]
	\caption{Ablation study of texture and motion branches.}
	\label{table:table7}
	\centering
	\resizebox*{\columnwidth}{!}{
		\begin{tabular}{l|ccc}
			\toprule[0.15em]
            Model & Vid4~\cite{liu2013bayesian} & REDS4~\cite{nah2019ntire} & Params(M) \\ 
                \midrule
                Base & 27.44/0.8284 & 31.58/0.8932 & 6.29 \\
                Base + Texture & 29.51/0.8909 & 32.79/0.9174 & 8.90 \\
                Base + Motion & 28.22/0.8647 & 32.43/0.9043 & 8.94 \\
                Base + Texture + Motion & 29.78/0.8983 & 32.93/0.9195 & 10.15 \\
			\bottomrule[0.15em] 
		\end{tabular}
	}
\end{table}
    
    \begin{table}[t!]
	\caption{Ablation studies about important factors of the Iterative Texture Enhancement module on Vid4~\cite{liu2013bayesian}.}
	\label{table:table8}
	\centering
	\resizebox*{\columnwidth}{!}{
		\begin{tabular}{ccccccc}
			\toprule
			\multirow{2}[1]{*}{\makecell{Model \\ ID}} & \multirow{2}[1]{*}{\makecell{Texture \\ Updater}}  & \multirow{2}[1]{*}{\makecell{Iterative \\ Manner}} & \multirow{2}[1]{*}{\makecell{Residual \\ Learning}} & \multirow{2}[1]{*}{\makecell{Iteration \\ Number}} & \multirow{2}[1]{*}{\makecell{\#Params \\ (M)}} & \multirow{2}[1]{*}{\makecell{PSNR}}\\ [0.2em]
			 &  &  & & &  &  \\
			\midrule
            (a) & Conv & \Checkmark & \Checkmark & 5 & 8.8 & 29.17 \\
			(b) & - & \XSolidBrush & \Checkmark & - & 7.6 & 29.09 \\
			(c) & ConvGRU & \Checkmark & \XSolidBrush & 5 & 8.5 & 29.16 \\
			\midrule
			(d) & ConvGRU & \Checkmark & \Checkmark & 3 & 8.9 & 29.46 \\
			(e) & ConvGRU & \Checkmark & \Checkmark & 8 & 8.9 & 29.32 \\
			Ours & ConvGRU & \Checkmark & \Checkmark & 5 & 8.9 & \textbf{29.51} \\
			\bottomrule
		\end{tabular}
	}
\end{table}

\subsubsection{Effect of Iterative Texture Enhancement}

We further analyze the proposed ITE module. To decouple its contribution, we remove the event-guided motion branch during this ablation study. The results are reported in Tab.~\ref{table:table8}. The specific configurations of each variant are detailed as follows.

Model (a) substitutes the ConvGRU block in the texture updater with standard convolutional layers. While retaining the iterative structure and residual learning, it lacks the explicit hidden state updating mechanism, leading to a performance drop. Model (b) removes the iterative refinement mechanism entirely. Instead of processing event bins sequentially, we employ a U-Net to extract texture features from the entire event voxel grid in a single pass. This non-iterative approach results in a 0.42 dB PSNR drop, highlighting the importance of temporal recurrence. Model (c) evaluates the residual connection by removing the element-wise addition in Eqs.~(\ref{eq6}) and~(\ref{eq7}). Forcing the network to directly predict the target feature $f_t^T$ rather than refining the input via a residual $\Delta$ leads to a 0.35 dB PSNR drop, confirming that learning high-frequency residuals is more effective for texture recovery. Finally, models (d-e) vary the number of iterations. Results show that 5 iterations yield the best performance. Increasing to $N=8$ provides no further gain, likely because over-segmenting the event stream results in sparse representations, degrading feature extraction.

\begin{figure}[t!]
        \centering
        \includegraphics[width=\columnwidth]{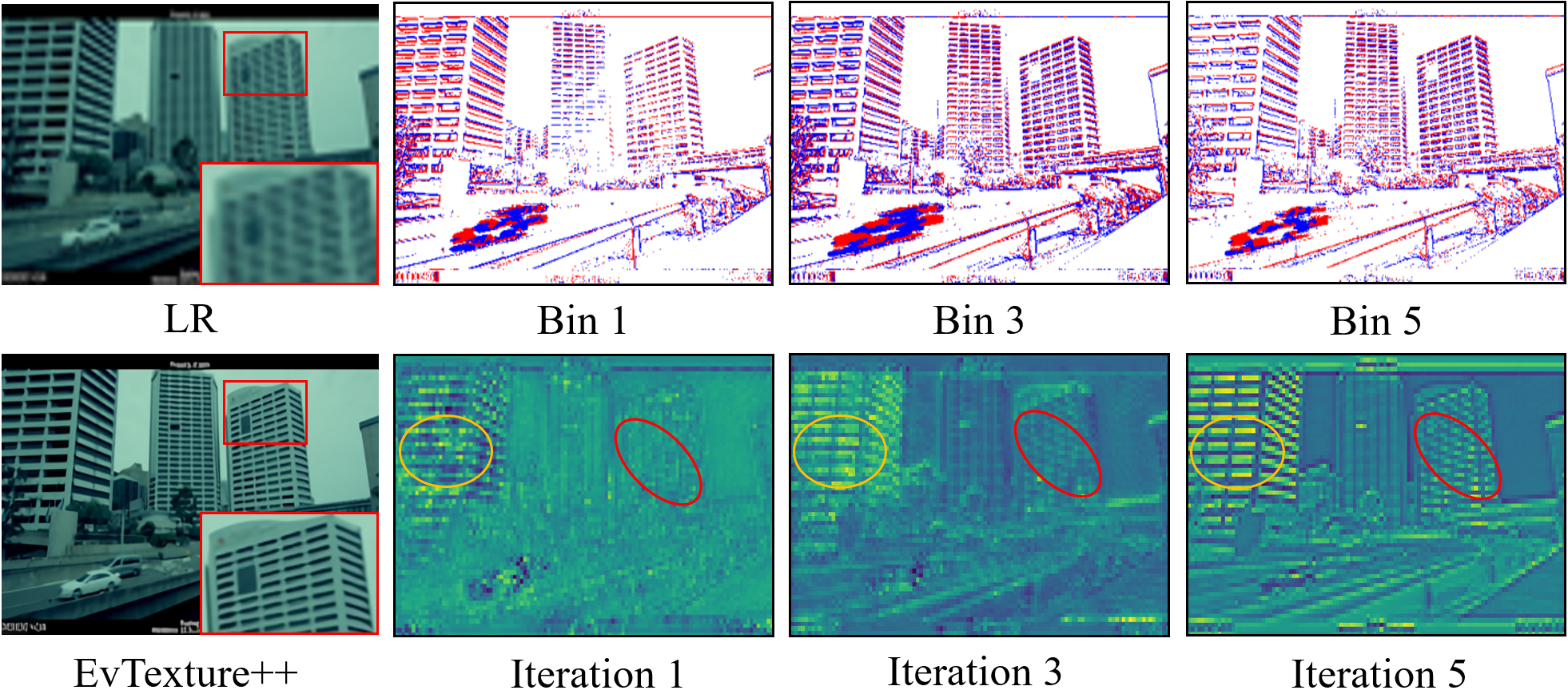}
        \caption{Visual results of the iterative texture enhancement process. As the iterations advance, the intermediate features capture clearer textural details from the event voxel grids, progressively enhancing the restoration quality.}
        \label{fig:fig19}
    \end{figure}

Additionally, we visualize the progression of intermediate features in Fig.~\ref{fig:fig19}. As the texture updater advances, EvTexture++ can progressively learn to extract finer textures with less noise and more clarity from voxel bins for restoration.

\section{Discussion} \label{sec:discussion}
In this section, we analyze the effectiveness of texture restoration (Sec.~\ref{sec:discuss_texture}), evaluate the robustness of our method under large motion and BD degradation (Sec.~\ref{sec:discuss_robustness}), compare against concurrent works (Sec.~\ref{discuss:Concurrent}), and discuss applicability to real-world event data (Sec.~\ref{disc:sim_vs_real}).

    \begin{figure}[t!]
        \centering
        \includegraphics[width=\columnwidth]{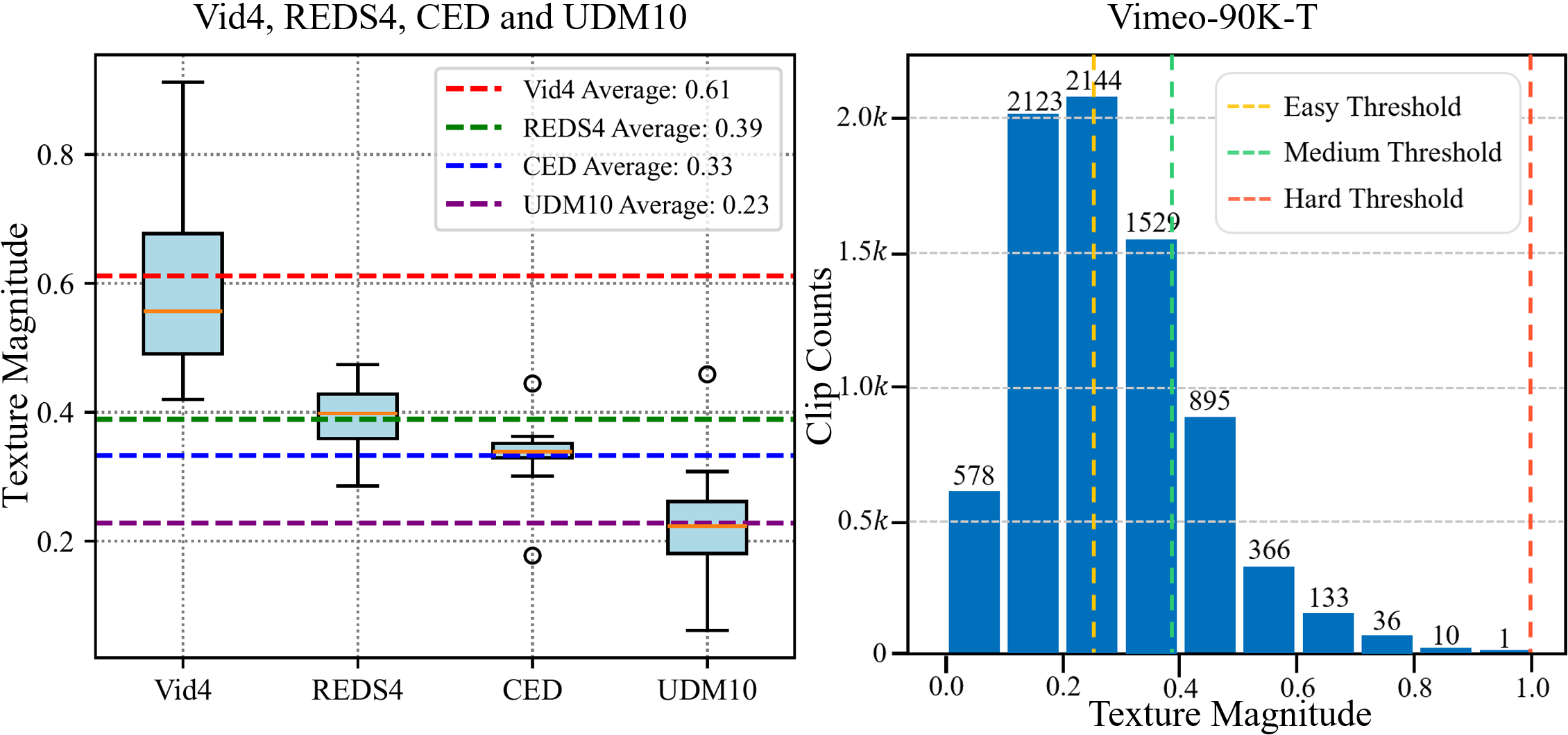}
        \caption{Texture magnitude analysis of five datasets. The results reveal that Vid4 has the most significant texture. Vimeo-90K-T covers a wide range and is divided into three difficulty levels: easy, medium, and hard.}
        \label{fig:fig20}
    \end{figure}

\subsection{Effectiveness of Texture Restoration} \label{sec:discuss_texture}

To quantify the effectiveness of our method in texture restoration, drawing inspiration from texture analysis in SISR~\cite{cai2022tdpn}, we first compute the texture magnitude of a video clip. Given a ground-truth video with $T$ frames of size $H \times W$, the texture magnitude is defined as:
\begin{equation}\label{eq:eq16}
    \frac{\alpha}{T}\sum_{t=1}^{T}\sqrt{\frac{1}{HW}\sum_{i=1}^{H}\sum_{j=1}^{W}\left|I_t(i,j) - \bar{I}_t(i,j)\right|^2}.
\end{equation}
Here, each frame $I$ is smoothed using a Gaussian filter to obtain the blurred image $\bar{I}$, with a kernel size of $(5, 5)$ and $\sigma=1.5$. We calculate the absolute difference between the original and blurred images to extract high-frequency details, averaging the contrast across the sequence. $\alpha$ is a scaling factor set to 10. The texture magnitude in Eq.~(\ref{eq:eq16}) ranges from 0 to 1, where a higher value indicates richer texture complexity.

    \begin{table*}[h!]
    \caption{Clip-by-clip results (PSNR$\uparrow$/SSIM$\uparrow$) on REDS4~\cite{nah2019ntire} for 4$\times$ VSR. Texture magnitude (last column) reflects texture complexity. Values marked with a \uwave{wavy underline} indicate larger gains on texture-rich clips compared to texture-poor ones.}
	\label{table:table9}
	\centering
	\resizebox{\textwidth}{!}{
	\begin{tabular}{lccccccccc} 
		\toprule[0.15em]
		\multirow{2}[2]{*}{\makecell{Clip Name}} & \multicolumn{3}{c}{RGB-based VSR} & \multicolumn{3}{c}{Event-based VSR} & \multirow{2}[2]{*}{\makecell{ EvTexture++\\\textit{vs.} VRT}} & \multirow{2}[2]{*}{\makecell{ EvTexture++\\\textit{vs.} EBVSR}} & \multirow{2}[2]{*}{\makecell{Tex. Mag. \\ (Eq.~(\ref{eq:eq16}))}} \\
		\cmidrule(lr){2-4} \cmidrule(lr){5-7} 
		& BasicVSR~\cite{chan2021basicvsr} & TTVSR~\cite{liu2022learning} &  VRT~\cite{liang2024vrt} & EGVSR~\cite{lu2023learning} & EBVSR~\cite{kai2023video} & \textbf{EvTexture++} & & \\
  
		\midrule
		Clip\_000 & 28.40/0.8434 & 28.82/0.8565 & 28.85/0.8553 & 25.16/0.7066 & 28.44/0.8446 & \textbf{30.90}/\textbf{0.9116} & \uwave{+2.05/+0.0563} & \uwave{+2.46/+0.0670} & \uwave{0.47} \\
  
		Clip\_011 & 32.47/0.8979 & 33.46/0.9100 & 33.49/0.9072 & 26.56/0.7722 & 32.55/0.8987 & \textbf{33.90}/\textbf{0.9168} & +0.41/+0.0096 & +1.35/+0.0181 & 0.38 \\
  
		Clip\_015 & 34.18/0.9224 & 35.01/0.9325 & \textbf{35.26}/\textbf{0.9332} & 29.83/0.8526 & 34.22/0.9235 & 35.14/0.9326 & \uwave{-0.12/-0.0006} & \uwave{+1.02/+0.0091} & \uwave{0.29} \\
  
		Clip\_020 & 30.63/0.9000 & 31.17/0.9093 & 31.16/0.9078 & 25.94/0.7846 & 30.67/0.9009 & \textbf{31.77}/\textbf{0.9170} & +0.61/+0.0092 &  +1.10/+0.0161 & 0.41 \\
  
		\midrule
		Average & 31.42/0.8909 & 32.12/0.9021 & 32.19/0.9006 & 26.87/0.7790 & 31.47/0.8919 & \textbf{32.93}/\textbf{0.9195}  & +0.74/+0.0189 &  +1.46/+0.0276 & 0.39 \\
		\bottomrule[0.15em]
	\end{tabular}
 	}
\end{table*}

    \begin{table}[t!]
    \caption{Quantitative comparison (PSNR$\uparrow$/SSIM$\uparrow$) across easy, medium, and hard difficulty levels of Vimeo-90K-T~\cite{xue2019video} for 4$\times$ VSR.}
    \label{table:table10}
    \centering
    \resizebox{\columnwidth}{!}{
        \begin{tabular}{ l|ccc}
            \toprule[0.15em]
            \multirow{2}[2]{*}{Method} & \multicolumn{3}{c}{Vimeo-90K-T~\cite{xue2019video}} \\ [-0.1em]
            \cmidrule(lr){2-4} 
            & Easy & Medium & Hard\\
            \midrule
            EDVR~\cite{wang2019edvr} & 41.98/0.9747 & 35.10/0.9422 & 30.40/0.8943 \\
            BasicVSR~\cite{chan2021basicvsr}  & 41.55/0.9731 & 34.63/0.9368 & 29.97/0.8864 \\
            IconVSR~\cite{chan2021basicvsr} & 41.73/0.9742 & 34.86/0.9387 & 30.19/0.8901 \\
            BasicVSR++~\cite{chan2022basicvsr++} & 41.98/0.9750 & 35.09/0.9412 & 30.38/0.8933 \\
            RVRT~\cite{liang2022recurrent} & 42.43/\underline{0.9762} & 35.69/0.9457 & 30.73/\underline{0.9022} \\
            VRT~\cite{liang2024vrt} & \textbf{42.47}/0.9761 & 35.73/\underline{0.9472} & \underline{30.74}/0.9018 \\
            EGVSR~\cite{lu2023learning} & 38.75/0.9594 & 32.15/0.9051 & 27.90/0.8358 \\
            EBVSR~\cite{kai2023video} & 41.55/0.9731 & 35.09/0.9412 & 30.49/0.8963 \\
            \textbf{EvTexture++} & \textbf{42.47}/\textbf{0.9784} & \textbf{35.85}/\textbf{0.9493} & \textbf{31.24}/\textbf{0.9104} \\
            \midrule
            $\#$ of clips & 3,907 & 2,345 & 1,563\\
            Avg. Texture Mag. & 0.16 & 0.32 & 0.49 \\
            \bottomrule[0.15em]
        \end{tabular}
    }
    \end{table}

We apply this metric to analyze five benchmark datasets: Vid4~\cite{liu2013bayesian}, REDS4~\cite{nah2019ntire}, CED~\cite{scheerlinck2019ced}, UDM10~\cite{yi2019progressive}, and Vimeo-90K-T~\cite{xue2019video}. The results in Fig.~\ref{fig:fig20} reveal Vid4 has the highest texture magnitude, followed by REDS4, CED, and UDM10. Correlating this with prior experimental results, we observe that our method performs particularly well on texture-rich datasets. For example, on Vid4 (rich texture) and CED (moderate texture), Tab.~\ref{table:table1} and Tab.~\ref{table:table3} demonstrate that EvTexture++ outperforms EBVSR~\cite{kai2023video} by 1.32 dB and 0.29 dB respectively, highlighting its greater effectiveness on complex textures.

We further analyze performance at the clip level on the REDS4~\cite{nah2019ntire} dataset. Tab.~\ref{table:table9} presents the results on four representative clips. On the most texture-rich clip, `Clip\_000', EvTexture++ shows substantial gains of 2.46 dB over EBVSR and 2.05 dB over VRT~\cite{liang2024vrt}. In contrast, the texture-poor `Clip\_015' yields a smaller improvement of 1.02 dB over EBVSR and performs comparably to VRT (-0.12 dB). This observation confirms that our method is specifically optimized for enhancing regions with rich high-frequency textures.

We also perform a detailed analysis on Vimeo-90K-T, which contains 7,824 clips with varying texture complexity. After removing 9 all-black clips~\cite{haris2019recurrent}, we divide the remaining 7,815 clips into easy, medium, and hard subsets based on their texture magnitude (Fig.~\ref{fig:fig20}). As presented in Tab.~\ref{table:table10}, EvTexture++ achieves comparable results to VRT on the easy subset, while on the hard subset, it attains gains of up to 0.50 dB. This further indicates that the performance improvement of our method is positively correlated with texture complexity.

While quantifying texture magnitude in VSR remains an open problem, our proposed metric offers a feasible preliminary solution. The evaluation results demonstrate its utility in characterizing dataset complexity and validating the texture-specific advantages of EvTexture++.

\subsection{Robustness Analysis} \label{sec:discuss_robustness}

\subsubsection{Robustness to Large Motion} \label{sec:discuss_motion}

To assess the robustness of EvTexture++ under large motion, we examine its performance relative to motion magnitudes. Specifically, we compare the motion magnitude distributions of REDS4~\cite{nah2019ntire} and CED~\cite{scheerlinck2019ced}, and analyze whether our method delivers larger gains in high-motion scenarios.

\begin{figure}[t!]
    \centering
    \includegraphics[width=\linewidth]{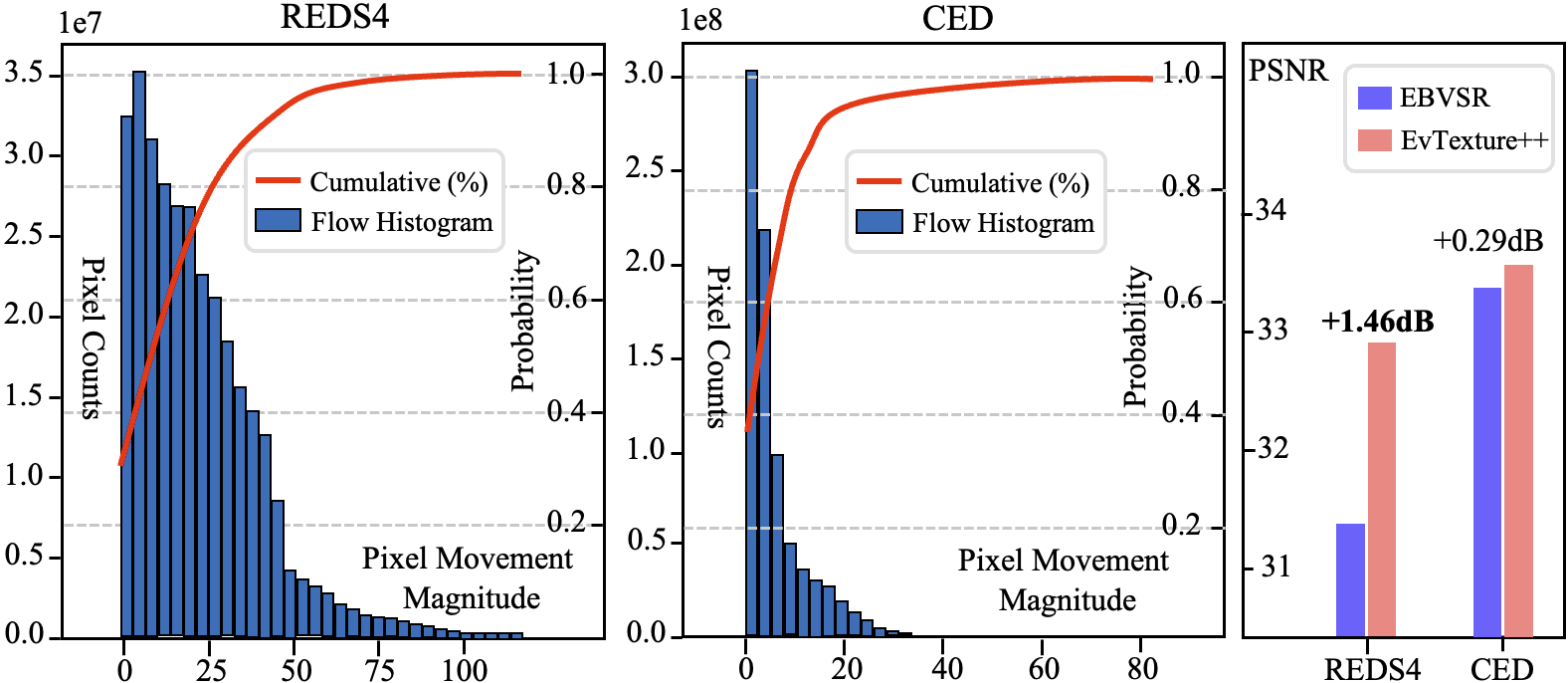}
    \caption{Pixel movement distributions of the REDS4~\cite{nah2019ntire} and CED~\cite{scheerlinck2019ced} test sets. REDS4 shows larger motion, where EvTexture++ achieves higher gains (+1.46 dB) over EBVSR~\cite{kai2023video}, indicating better robustness to large motion.}
    \label{fig:fig21}
\end{figure}

We compute the optical flow between adjacent ground-truth frames using RAFT~\cite{teed2020raft} and calculate the per-pixel flow magnitude $\sqrt{u^2 + v^2}$. Fig.~\ref{fig:fig21} depicts the distribution of pixel counts over different movement magnitudes for the test sets. The distribution exhibits a long-tailed pattern. Notably, REDS4 contains significantly more pixels with magnitudes exceeding 50, while in CED, most motions are below 20 pixels. These results indicate that REDS4 involves substantially larger motion overall. Comparing EvTexture++ against EBVSR~\cite{kai2023video}, we observe that on the high-motion REDS4, EvTexture++ achieves a gain of 1.46~dB, whereas on the lower-motion CED, the gain is 0.29~dB. These results suggest that our method offers enhanced robustness and larger improvements in scenarios characterized by large motion.

\begin{table}[t!]
    \centering
    \caption{Quantitative comparison (PSNR$\uparrow$/SSIM$\uparrow$) under BD degradation ($4\times$). $^{\dagger\dagger}$ denotes models enhanced by our plug-in \colorbox{customgray}{EvTexture++}.}
    \label{table:table11}
    \resizebox{\columnwidth}{!}{
        \begin{tabular}{lccc}
            \toprule[0.15em]
            \multirow{2}[2]{*}{\makecell{Method}} & \multicolumn{3}{c}{BD Degradation}\\
            \cmidrule(lr){2-4}
            & UDM10~\cite{yi2019progressive} & Vimeo-90K-T~\cite{xue2019video} & Vid4~\cite{liu2013bayesian} \\
            \midrule
            EDVR~\cite{wang2019edvr} & 39.89/0.9686  & 37.81/0.9523 & 27.85/0.8503 \\
            BasicVSR~\cite{chan2021basicvsr} & 39.96/0.9694 & 37.53/0.9498 & 27.96/0.8553 \\
            IconVSR~\cite{chan2021basicvsr} & 40.03/0.9694 & 37.84/0.9524 & 28.04/0.8570 \\
            BasicVSR++~\cite{chan2022basicvsr++} & 40.72/\underline{0.9722} & 38.21/0.9550 & 29.04/0.8753 \\
            MIA-VSR~\cite{zhou2024video} & 39.35/0.9668 & 37.09/0.9463 & 27.63/0.8430 \\
            IART~\cite{xu2024enhancing} & \underline{41.15}/\textbf{0.9750} & \underline{38.63}/\underline{0.9582} & \underline{29.60}/\underline{0.8868} \\
                        
            EBVSR~\cite{kai2023video} & 40.12/0.9704 & 37.64/0.9508 & 28.26/0.8641\\
            EvTexture~\cite{kai2024evtexture} & 40.27/0.9705 & 37.80/0.9517 & 28.91/0.8744 \\

            \midrule
            \textbf{EvTexture++} & 40.53/\underline{0.9722} & 38.02/0.9540 & 29.24/0.8865 \\
            
            \rowcolor{customgray}
            \textbf{IART$^{\dagger\dagger}$} & \textbf{41.27}/\textbf{0.9750} & \textbf{38.82}/\textbf{0.9592} & \textbf{29.86}/\textbf{0.8893} \\
            
            \bottomrule[0.15em]
        \end{tabular}
    }
\end{table}

\begin{table}[t!]
    \centering
    \caption{Quantitative comparison with concurrent event-driven VSR methods on the Vid4~\cite{liu2013bayesian} dataset for 4$\times$ VSR.}
    \label{table:table12}
    
    \resizebox{1.0\columnwidth}{!}{
        \begin{tabular}{lcccc}
            \toprule[0.15em]

            \multirow{2}[1]{*}{\makecell{Metric}} & \multirow{2}[1]{*}{\makecell{MamEVSR\\\cite{Xiao_2025_CVPR}}} & \multirow{2}[1]{*}{\makecell{EvSTVSR\\\cite{yan2025evstvsr}}} & \multirow{2}[1]{*}{\makecell{EvEnhancer\\\cite{wei2025evenhancer}}} & \multirow{2}[1]{*}{\makecell{\textbf{EvTexture++}}}\\[1.3em]
            
            \midrule
            
            PSNR$\uparrow$ & 28.52 & 28.03 & 28.42 &  \textbf{29.78} \\
            SSIM$\uparrow$ & 0.8804 & 0.8309 & 0.8774 &  \textbf{0.8983} \\
            LPIPS$\downarrow$ & 0.2372 & 0.2543 & 0.2401 &  \textbf{0.2048} \\
            
            \midrule
            
            TCC$\uparrow$\scalebox{0.5}{$ \times  10$} & 3.037 & 2.774 & 2.972 &  \textbf{3.832} \\
            tOF$\downarrow$\scalebox{0.5}{$ \times  10$} & 1.528 & 1.532 & 1.688 &  \textbf{1.158} \\
            
            \midrule
            
            \#Params(M) & 9.58 & 25.62 & \textbf{6.55} &  10.15 \\
            Runtime(ms) & 145.2 & 567.7 & 1240.4 &  \textbf{95.0} \\
            FLOPs(G) & \textbf{623.7} & 1430.9 & 1356.3 &  808.6 \\
            
            \bottomrule[0.15em]
        \end{tabular}
    }
\end{table}

\subsubsection{Performance under BD Degradation} \label{disc:bd_degradation}

We further evaluate robustness under BD degradation. This setting heavily suppresses high-frequency spatial details. Quantitative comparisons are reported in Tab.~\ref{table:table11}. As shown, EvTexture++ demonstrates exceptional robustness. Specifically, it surpasses its backbone, BasicVSR~\cite{chan2021basicvsr}, by a substantial margin of 1.28 dB on Vid4. It also consistently outperforms the representative event-based method, EBVSR~\cite{kai2023video}. Moreover, to verify generalizability, we apply our plug-in to the advanced Transformer-based model, IART~\cite{xu2024enhancing}. While IART is naturally strong, enhancing it with our module further boosts performance to state-of-the-art levels (\textit{e.g.}, +0.26 dB on Vid4). This confirms that utilizing event signals remains highly effective for recovering lost details.

\subsection{Comparison with Concurrent Works} \label{discuss:Concurrent}

We also compare EvTexture++ with three concurrent event-driven VSR methods: MamEVSR~\cite{Xiao_2025_CVPR}, EvSTVSR~\cite{yan2025evstvsr}, and EvEnhancer~\cite{wei2025evenhancer}. As official implementations for some methods are unavailable, we carefully reproduced these methods following their original papers. To ensure fairness, all methods were trained and evaluated under identical settings.

The comparative results on the texture-rich Vid4~\cite{liu2013bayesian} dataset are summarized in Tab.~\ref{table:table12}. As observed, EvTexture++ consistently outperforms these concurrent approaches across all quality metrics. Specifically, in terms of spatial restoration, our method surpasses the second-best method, MamEVSR, by 1.26 dB in PSNR and achieves the lowest LPIPS score (0.2048), indicating superior capability in recovering fine-grained textures. Regarding temporal stability, EvTexture++ attains the best TCC and tOF scores, validating the effectiveness of the proposed TTA module in suppressing flickering artifacts. Furthermore, concerning computational efficiency, although EvEnhancer has the fewest parameters, its inference latency is notably high ($>$1200ms). In contrast, EvTexture++ maintains the most competitive balance between performance and cost, achieving the fastest runtime (95.0ms) among all compared methods while delivering SOTA restoration quality.

\subsection{Analysis of Sim-to-Real Domain Gap} \label{disc:sim_vs_real}

Most event-driven VSR benchmarks~\cite{liu2013bayesian},~\cite{xue2019video},~\cite{nah2019ntire} rely on the ESIM simulator~\cite{rebecq2018esim} due to the scarcity of high-quality ground truth for real event cameras. While synthetic data facilitates controlled training, we acknowledge the inherent domain gap between simulation and reality. Real-world event sensors (\textit{e.g.}, DAVIS346~\cite{brandli2014real} or Prophesee) introduce non-ideal characteristics, such as stochastic background noise (hot pixels), temporal jitter, and threshold variations, which are absent in ideal simulation. Consequently, directly applying a model trained on synthetic data to real-world scenarios often degrades performance, as the network may misinterpret transient sensor noise as high-frequency textural details.

To address these challenges, we empirically analyze the architectural robustness of EvTexture++ on the CED~\cite{scheerlinck2019ced} dataset (see Tab.~\ref{table:table3}). By training and evaluating on real-world event streams, our method achieves SOTA performance (33.71 dB PSNR) on this benchmark. We attribute this robustness to the design of the ITE module. The internal ConvGRU mechanism exploits temporal consistency: since valid textural structures are temporally stable while sensor noise is typically random and transient, the recurrent unit acts as a temporal filter. This mechanism suppresses noise while accumulating valid texture cues, ensuring that our framework adapts effectively to the complex real-world sensor characteristics.

\section{Conclusion} \label{sec:conclusion}

This paper presents EvTexture++, an event-driven framework dedicated to texture restoration in VSR. The framework adopts a two-branch structure, comprising an event-driven texture branch and a motion branch. In the texture branch, the ITE module progressively refines textures using high-temporal-resolution events, while in the motion branch, the TTA module enhances temporal consistency by leveraging complementary motion information from both events and RGB frames. In addition, EvTexture++ can serve as a plug-and-play module for existing VSR backbones, delivering substantial performance gains with minimal overhead. Extensive experiments verify its superior performance, particularly in texture-rich scenarios.

\textbf{Limitations and Future Work.} Our work has two limitations. First, we assume precise spatial alignment and identical resolution between event and RGB inputs. However, such a configuration is limited to a few specialized sensors. In practical systems, spatial parallax often exists, and event cameras typically have lower resolution. Therefore, real-world applications should account for these physical discrepancies. Second, our current evaluation is limited to deterministic CNN- and Transformer-based backbones. While extending EvTexture++ to diffusion-based VSR models~\cite{yang2024motion},~\cite{wang2025seedvr} is a promising direction, integrating events into the stochastic diffusion process involves higher computational costs and complex training strategies. We leave this for future exploration.

\normalem
\bibliographystyle{IEEEtran}
\bibliography{reference}

\begin{IEEEbiography}[{\includegraphics[width=1in,height=1.25in,clip,keepaspectratio]{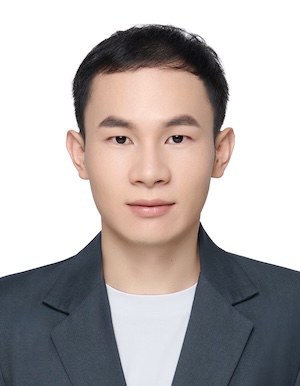}}]{Dachun Kai}
received the B.S. degree from Hefei University of Technology (HFUT), Hefei, China, in 2021. He is currently pursuing the Ph.D. degree with the University of Science and Technology of China (USTC), Hefei, China. His research interests include computational photography, image and video processing, and deep learning.
\end{IEEEbiography}

\begin{IEEEbiography}[{\includegraphics[width=1in,height=1.25in,clip,keepaspectratio]{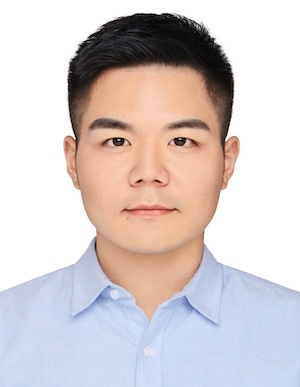}}]{Jiayao Lu}
received the B.S. degree from University of Science and Technology of China (USTC), Hefei, China, in 2018. He is currently pursuing a Ph.D. degree at the USTC. His research interests include video understanding, computer vision, and deep learning.
\end{IEEEbiography}

\begin{IEEEbiography}[{\includegraphics[width=1in,height=1.25in,clip,keepaspectratio]{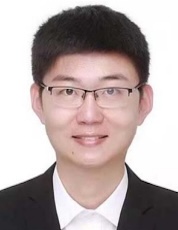}}]{Yueyi Zhang}
(Member, IEEE) received the B.S. degree in applied mathematics and the Ph.D. degree in information and communication engineering from the University of Science and Technology of China (USTC), Hefei, China, in 2010 and 2015, respectively. He is currently with Midea Group, Shanghai, China. Prior to this, he was an Associate Researcher with USTC. He has authored or co-authored more than 90 papers in refereed journals and conferences. His research interests include computational photography and event-based vision.
\end{IEEEbiography}

\begin{IEEEbiography}[{\includegraphics[width=1in,height=1.25in,clip,keepaspectratio]{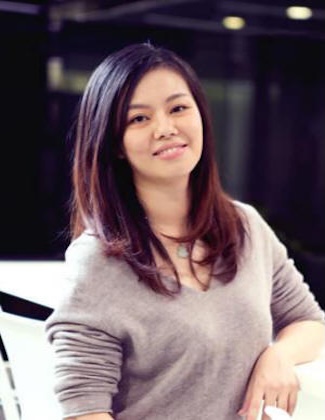}}]{Xiaoyan Sun}
(Senior Member, IEEE) received the B.S., M.S., and Ph.D. degrees in computer science from Harbin Institute of Technology, Harbin, China, in 1997, 1999, and 2003, respectively. 

She is currently a Full Professor with the University of Science and Technology of China, Hefei, China, and the Deputy Director of the National Engineering Laboratory for Brain-inspired Intelligence Technology and Application. Prior to joining USTC in 2019, she was with Microsoft Research Asia, Beijing, China, from 2003 to 2019, most recently as a Senior Researcher. Since 2000, she has been an Intern with Microsoft Research Asia before joining Microsoft Research Asia in 2003. She has authored or co-authored more than 100 publications in journals and conferences, and contributed to ten standard proposals (with one accepted). She holds more than ten granted U.S. patents. Her research interests include computer vision, image/video processing, machine learning, and artificial intelligence. 

Dr. Sun was a recipient of the Best Paper Award of IEEE TRANSACTIONS ON CIRCUITS AND SYSTEMS FOR VIDEO TECHNOLOGY in 2009 and the Best Student Paper Award of VCIP 2016. She serves on the Senior Editorial Board of IEEE JOURNAL ON EMERGING AND SELECTED TOPICS IN CIRCUITS AND SYSTEMS and as an Associate Editor for \textit{Signal Processing: Image Communication}.

\end{IEEEbiography}

\end{document}